\documentclass[journal,10pt]{IEEEtran} 
\usepackage{balance}
\usepackage[ruled,vlined,linesnumbered,resetcount]{algorithm2e}
\SetKwRepeat{Do}{do}{while}%
\usepackage[linesnumbered,ruled,vlined]{algorithm2e}
\usepackage[sort&compress,numbers]{natbib}
\usepackage{amsmath,amssymb,amsfonts,bm}

\usepackage{textcomp}
\usepackage{graphicx} 
\usepackage{subfigure}
\usepackage{booktabs}
\usepackage{multirow}
\usepackage{epsfig}
\usepackage{amssymb}
\usepackage{amsthm}
\usepackage{times} 
\usepackage{amsmath} 
\usepackage{amssymb}  
\usepackage{mathrsfs}
\usepackage{epstopdf}
\usepackage{array}
\usepackage{xcolor}    
\usepackage{colortbl}  
\usepackage{color}
\usepackage[numbers]{natbib} 
\usepackage{mathtools}

\def\BibTeX{{\rm B\kern-.05em{\sc i\kern-.025em b}\kern-.08em
    T\kern-.1667em\lower.7ex\hbox{E}\kern-.125emX}}
\usepackage{url}
\usepackage{tikz} 
\usetikzlibrary{shapes,shapes.geometric,arrows,fit,calc,positioning,automata,}

\renewcommand{\SetKwInOut}[2]{%
	\sbox\algocf@inoutbox{\KwSty{#2}\algocf@typo:}%
	\expandafter\ifx\csname InOutSizeDefined\endcsname\relax
	\newcommand\InOutSizeDefined{}%
	\sbox\algocf@inoutbox{\KwSty{#2}\algocf@typo\textbf{:}~}\setlength{\inoutindent}{\wd\algocf@inoutbox}%
	\else
	\ifdim\wd\algocf@inoutbox>\inoutsize%
	\sbox\algocf@inoutbox{\KwSty{#2}\algocf@typo\textbf{:}~}\setlength{\inoutindent}{\wd\algocf@inoutbox}%
	\fi%
	\fi
	\algocf@newcommand{#1}[1]{%
		\ifthenelse{\boolean{algocf@inoutnumbered}}{\relax}{\everypar={\relax}}%
		{\let\\\algocf@newinout\hangindent=\inoutindent\hangafter=1\KwSty{#2}\algocf@typo\textbf{:}~##1\par}%
		\algocf@linesnumbered
}}
\usepackage{titlesec}
\definecolor{headerbg}{RGB}{222, 216, 200}
\definecolor{colheaderbg}{RGB}{240, 237, 220}
\definecolor{rowgray}{gray}{0.95}

\begin{document}
\title{Koopman-Based Generalizability Analysis of Deep Reinforcement Learning With Application to Wireless Communications} 
\author{Atefeh Termehchi, Ekram Hossain,  \IEEEmembership{Fellow, IEEE}, and  Isaac Woungang,  \IEEEmembership{Senior Member, IEEE}
\thanks{Atefeh Termehchi and Ekram Hossain are with the Department of Electrical and Computer Engineering at the University of Manitoba, Winnipeg, Canada (emails: atefeh.termehchi@umanitoba.ca and ekram.hossain@umanitoba.ca). Isaac Woungang is with the Department of Computer Science, Toronto Metropolitan University, Toronto, Canada (email: iwoungan@torontomu.ca).}}
\maketitle
\begin{abstract}
Deep Reinforcement Learning (DRL) is a key machine learning technology driving progress across various scientific and engineering domains, including wireless communications. However, its limited interpretability and generalizability remain major challenges. In supervised learning, generalizability is often assessed through generalization error using information-theoretic methods, which typically assume that the training data is independent and identically distributed (i.i.d.). In contrast, DRL involves sequential and dependent data, rendering standard information-theoretic approaches unsuitable for analyzing its generalization performance. To address this, our work introduces a novel analytical method to evaluate the generalization of DRL. The particular focus is on the ability of DRL algorithms to generalize beyond the training domain. We approach this by developing a statistical understanding of the internal state dynamics of DRL algorithms.  Specifically, we model the evolution of states and actions in trained DRL algorithms as unknown, discrete, stochastic, and nonlinear dynamical systems. Domain changes are modeled as additive disturbance vectors affecting state evolution. To identify the underlying unknown nonlinear dynamics, we apply Koopman operator theory, enabling interpretable representations of the state-action evolution. Based on the interpretable representations, we perform spectral analysis using the H$_\infty$ norm to estimate the worst-case impact of domain changes on DRL performance. Finally, we apply our analytical framework to assess the generalizability of different DRL algorithms in a wireless communication environment.
\end{abstract}
\begin{IEEEkeywords}
  Generalizability, interpretability, deep reinforcement learning, Koopman operator, H$_\infty$ norm.
\end{IEEEkeywords}
\section{Introduction}
Many real-world problems across scientific and engineering fields, such as robotics, wireless communications and networking, involve complicated optimization problems. For instance, in modern wireless networks (e.g., 5G and 6G systems), tasks such as user association, resource allocation, multiple-antenna beamforming for joint communication and sensing, and joint active and passive beamforming in RIS-aided systems, require solving NP-hard problems. 
Traditional optimization methods, such as branch and bound, dynamic programming, and heuristics, can provide solutions. However, these solutions are often computationally expensive and impractical for large-scale dynamic environment. Model-free deep reinforcement learning (DRL) offers a promising alternative for decision-making in such environments \cite{sutton2018reinforcement, hoang2023deep}. Indeed, DRL can efficiently handle complex, high-dimensional optimization problems, making it a valuable tool across various fields of study. 

Despite its advantages over traditional optimization methods, DRL has two significant drawbacks: limited interpretability and generalizability \cite{karniadakis2021physics, glanois2024survey}. Interpretability refers to the model's ability to provide a clear, evidence-based explanation for a DRL decision. Specifically, it addresses the question: ``Why did the learning model decide that?" \cite{papernot2018deep}. Some studies \cite{glanois2024survey} distinguish interpretability as an intrinsic property and explainability as a post-hoc process. However, in this work, we consider them to be closely related and do not make a strict distinction. Generalizability refers to a model's ability to maintain good performance not only on the training data but also on unseen data. The ability to analyze generalizability is closely linked to the challenge of interpretability. A clear explanation for a model's decision can simplify assessing its performance on new data. 

In supervised learning, the generalization error is defined as the difference between the population risk, i.e., the expected value of the loss function over the true data distribution, and the empirical risk, i.e., the expected value of the loss function over the training set.
This error is the traditional metric to measure generalizability. Conventional methods for analyzing generalization error fall into two main categories: hypothesis class complexity-based bounds and information-theoretic bounds \cite{perlaza2024generalization}. Complexity-based methods, such as the Vapnik–Chervonenkis dimension and Rademacher complexity, assume that all models are equally likely. However, this assumption fails to capture data-dependent generalization, especially in modern deep neural networks (DNNs) \cite{hellstrom2025generalization}. In contrast,
information-theoretic approaches utilize measures like mutual information (MI) and the probably approximately correct (PAC)-Bayesian framework. These approaches quantify the dependence between the learned model and the training data.

However, employing these approaches in DNNs with millions of parameters requires integration over high-dimensional parameter spaces, making direct computation infeasible. 
Applying information-theoretic methods to generalizability analysis in DRL presents an additional challenge. These methods generally assume that training data is independent and identically distributed (i.i.d.). However, DRL collects observations by taking observation-dependent actions in an environment. In other words, DRL learns through interaction, with sequential training data that depend on past actions, making it non-independent training data sets. An emerging line of research aims to characterize generalization error in such interactive settings, including online learning and DRL, where data dependencies and anytime-valid results are critical. While some progress has been made, the field is still in its early stages with many open challenges \cite{rodriguez2024information}.

In addition, in theoretical machine learning (ML), the population risk is typically defined based on a distribution identical to that of the training data \cite{hellstrom2025generalization}. However, in many practical applications, models encounter distribution shifts after deployment. This challenge is known as out-of-distribution (OOD) generalization, which describes how effectively a model can apply its learned knowledge to make accurate predictions or decisions in new environments. In DRL, ensuring convergence to an optimal policy requires the assumption that the conditional transition probabilities of the underlying Markov decision process (MDP) remain stationary. As a result, understanding and addressing OOD generalization is even more crucial in DRL than in traditional supervised learning. This makes the analytical study of generalization in DRL both essential and complex.

The primary objective of this paper is to introduce an analytical method for evaluating the OOD generalizability of DRL algorithms. We begin by modeling the evolution of states and actions in trained DRL algorithms as unknown, discrete, stochastic, and nonlinear dynamical functions. Domain changes are then represented by incorporating an additive disturbance vector into the conditional transition probability function. To analyze the unknown dynamical functions, we employ Koopman operator theory, leveraging its data-driven identification capabilities. To approximate the spectral features of the Koopman operator, we apply both dynamic mode decomposition (DMD) and exact DMD. Subsequently, we use the 
 H$_\infty$ norm to assess these spectral characteristics and quantify the worst-case impact of domain changes on the trained DRL model.
\subsection{Motivation and Prior Work}
Several research efforts have explored DRL methods, including value-based and policy-based approaches (both deterministic and stochastic), to address various challenges in 5G and 6G wireless networks. However, as previously discussed, DRL techniques face two major challenges: limited interpretability and limited generalizability. Interpretability is particularly important in wireless applications, where reliability and safety are critical. Without a clear understanding of DRL-based decisions, it is difficult to ensure trustworthy system behavior. Moreover, limited generalizability can significantly hinder the effectiveness of DRL in dynamic and non-stationary wireless environments. To address these issues, the wireless communications community has recently focused on improving the generalization capabilities of DRL algorithms (see \cite{akrout2023domain} and references therein). Techniques such as transfer learning and domain adaptation have been proposed. However, these methods are not always practical. Fine-tuning or adaptation can introduce significant delays, which are often incompatible with the real-time requirements of wireless applications \cite{akrout2023domain}. Therefore, it is essential to rigorously study the interpretability and generalizability of DRL-based methods. Such efforts can guide the development of new learning algorithms and contribute to practical advancements. 
To the best of our knowledge, no prior work in wireless communication has analyzed the interpretability and generalizability of DRL using formal mathematical frameworks.

Over the past few decades, many studies within the ML community have focused on interpretability and generalizability. These topics remain active areas of research due to their significance and the numerous open challenges that persist \cite{ye2021towards,hellstrom2025generalization, zhou2022domain,rodriguez2024information, glanois2024survey,he2025information}.
Generalization error is a standard metric to analyze generalizability in supervised learning. As mentioned earlier, information-theoretic methods provide more practical insights into generalization behavior. However, applying information-theoretic generalization bounds, such as MI and PAC-Bayesian bounds, to deep learning (DL)-based methods presents significant challenges. For instance, MI requires knowledge of the true data distribution (i.e., the joint probability distribution of the input features and labels/outputs). Yet, this distribution is typically unknown in real-life applications, making exact MI computation impractical. Additionally, many information-theoretic metrics, such as Kullback–Leibler (KL) divergence and entropy, require integration over high-dimensional parameter spaces. Since DNNs often contain millions of parameters, computing these measures becomes infeasible. For example, PAC-Bayesian bounds face similar challenges, as they also require computing the KL divergence between the prior and posterior weight distributions. This computation often lacks closed-form solutions, resulting in high memory usage and computational costs. 

Furthermore, information-theoretic methods pose additional challenges in DRL. Unlike conventional DNN, where the training data is independent of the learning algorithm, DRL collects data through observation-dependent actions within an environment. As a result, the training data in DRL is sequential and dependent. Therefore, methods that assume i.i.d. data, such as those based on MI and PAC-Bayesian bounds, must be adapted to handle the sequential and dependent nature of training data in DRL \cite{rodriguez2024information, hellstrom2025generalization}. While some research has addressed this issue \cite{seldin2012pac, flynn2023pac, perlaza2024generalization}, the field remains in its early stages.

Moreover, traditional information-theoretic methods cannot be directly applied to analyze OOD generalization, as they typically evaluate population risk under the training distribution. In \cite{hellstrom2025generalization}, two methods are presented for analyzing OOD generalization bounds. The first is based on the KL divergence between the target and source (training) distributions. However, this method has strong limitations: it requires knowledge of both the training and target distributions, and more critically, it fails when these distributions have disjoint support. The second method uses the Wasserstein distance between the training and target distributions, which can handle the disjoint support scenario. Nevertheless, this approach still relies on access to both distributions, which is often impractical in real-world applications.

To understand the effect of domain change on the performance of DL algorithms, it is essential to study how the internal representations evolve under such changes. This involves statistically tracking the time step–evolving states in DRL or the intermediate feature maps at each layer in DNNs. 
Recent studies \cite{goldfeld2019estimating, shwartz2022information, he2025information} propose novel applications of MI to monitor the dynamics of intermediate feature maps across different layers. These approaches aim to make DNNs more interpretable and provide insights into their generalization behavior. In addition, Koopman operator theory and DMD have recently attracted attention for analyzing the dynamic behavior of internal states and parameters in DL–based approaches \cite{dogra2020optimizing, weissenbacher2022koopman, rozwood2024koopman}. Indeed, these two well-established, data-driven analytical tools offer promising directions for interpreting the black-box behavior of DL–based algorithms. In \cite{dogra2020optimizing}, the authors apply Koopman operator theory to predict the weights and biases of feedforward and fully connected DNNs during training. As a result, they report learning speeds over 10 times faster than conventional gradient descent–based optimizers such as Adam, Adadelta, and Adagrad. In a related study, \cite{rozwood2024koopman} demonstrates that the Koopman operator can capture the expected time evolution of a DRL value function dynamics. This ability enables the estimation of optimal value functions, ultimately enhancing the performance of the DRL algorithm.

\subsection{Contributions}
In this paper, we introduce a mathematical method to evaluate the OOD generalizability of DRL algorithms. The key contributions are as follows:
\begin{itemize}
\item We model the evolution of states and actions in trained DRL algorithms as unknown discrete dynamical stochastic nonlinear functions. In addition, we model domain changes over the conditional transition probability function of environments by an additive disturbance vector. 
\item We use the Koopman operator theory to identify the behavior of the unknown dynamical functions. Next, we employ DMD and exact DMD to approximate the spectral features of Koopman operator. Accordingly, we present two interpretable representations for the evolution associated with states and actions in the trained DRL algorithms.
\item Based on the approximated interpretable representations, we use the $Z$-transform and the H$_\infty$ norm, to quantify the maximum impact of domain changes on the trained DRL’s states and actions (see \textbf{Theorem 2} and \textbf{Corollary 1}). Then, we analyze the maximum effect of domain changes on the trained DRL performance in terms of the reward function  (see \textbf{Corollary 2}). 
\item  Based on \textbf{Theorem 2}, \textbf{Corollary 1}, and \textbf{Corollary 2}, we drive a bound on the generalization error for trained DRL algorithms (see \textbf{Corollary 3}). 
\end{itemize} 

\subsection{Organization and Notations}
The rest of this paper is organized as follows. In Section II, we provide the background, preliminaries, and definitions. In Section III, we model and identify the dynamical behavior of DRL. Section IV describes our proposed method for generalizability analysis in DRL. Finally, in Section V, the proposed method for generalizability analysis is applied to compare generalizability of DRL algorithms in 
a wireless communication scenario.

The following notations are used throughout this paper. The statistical expectation is represented by $\mathbb{E}$. For any given matrix $\mathbf{X}$, the element located at the $i$-th row and $j$-th column is denoted as $\mathbf{X}(i, j)$. The transpose and conjugate transpose of $\mathbf{X}$ are denoted by $\mathbf{X}^T$ and $\mathbf{X}^H$, respectively. The notation $\mathbf{x}_k$ refers to the vector $\mathbf{x}$ at time step $k$. The notation $\mathbf{x}_z$ denotes $Z$-transform version of the vector $\mathbf{x}$. The notation $\|\mathbf{x}\|$ is used for the norm of the vector $\mathbf{x}$. The absolute value of a number $x$ is written as $|x|$. The notation $\overline{\mathbf{x}}$ is used for the expected value of $\mathbf{x}$ over multiple independent realizations. Table \ref{tab: Table1_} provides a summary of the key notations used throughout the paper. 
\begin{table}[t!]
\centering
\renewcommand{\arraystretch}{1.4}
\caption{Table of notations}
\resizebox{\columnwidth}{!}{
\begin{tabular}{c|c}  
    \hline\hline
    \rowcolor{gray!20} \textbf{Parameters/Variables} &\textbf{Description} \\ 
    \hline
    $\mathcal{K}$ &Koopman operator \\
    \hline
    \rowcolor{gray!15} $\mathbf{\widetilde{K}}$ &Approximated Koopman operator \\
    \hline
    $\mathbf{x}$, $\mathbf{u}$  &State, Action \\
    \hline
    $k$, $\mathscr{K}$  &Time step, Set of time steps\\
    \hline
    \rowcolor{gray!15} $\mathbf{w}$  &Additive disturbance \\ 
    \hline
    $\bar{\mathbf{x}}^n$ &Expected value of state without domain change \\
    \hline
    \rowcolor{gray!15}  $\bar{\mathbf{x}}^{\text{w}}$  &Expected value of state in case of domain change \\
    \hline
    $\bar{\mathbf{u}}^n$ &Expected value of action without domain change \\
    \hline
    \rowcolor{gray!15}  $\bar{\mathbf{u}}^{\text{w}}$  &Expected value of action in case of domain change \\
    \hline
\end{tabular}
}
\label{tab: Table1_}
\end{table}
\section{Background, Preliminaries, and Definitions}
This section outlines the essential background theory, algorithm, and mathematical tools. Specifically, we discuss the definition of domain change and generalization error  in DRL, the Koopman operator theory, the DMD algorithm, and we provide a review of the $Z$-transform and the H$_\infty$ norm, which form the basis for the generalizability analysis presented in the following section.
\subsection{Definition of Domain Change in DRL}
Domain changes in supervised learning can generally be categorized into two types: covariate shift (also called input shift) and concept drift. The covariate shift occurs when the distribution of the input data changes between training and testing, while the conditional distribution of the output given the input remains the same~\cite{sugiyama2007covariate}. In contrast, concept drift refers to changes in the conditional distribution of the output given the input, even when the input distribution remains unchanged~\cite{gama2014survey}. This means the relationship between input and output changes over time.

In DRL, learning does not rely on an input-output dataset like in supervised learning. Instead, learning takes place through interaction with an environment. This environment is typically modeled by a conditional transition probability function, which defines the probability of transitioning to the next state given the current state and action. The goal in DRL is to learn an optimal action policy that maximizes expected rewards over time through this interaction. Therefore, domain change in DRL can be categorized into two main types:
\begin{itemize}
    \item Changes in the transition dynamics of environment, i.e., changes in the conditional transition probability function:
    \[
p_{\text{test}}(\mathbf{x}_{k+1} \mid \mathbf{x}_{k}, \mathbf{u}_{k}) \neq p_{\text{train}}(\mathbf{x}_{k+1} \mid \mathbf{x}_{k}, \mathbf{u}_{k}),
\]
where $\mathbf{x}_k \in \mathbb{R}^n$ is the state vector at time step $k$, and $\mathbf{u}_k \in \mathbb{R}^m$ is the action taken at time step $k$ according to the trained policy. 
    \item Changes in the reward function $r( \mathbf{x}_{k+1},\mathbf{u}_{k})$, which directly affect the learning objective:
    \[r_{\text{test}}( \mathbf{x}_{k+1},\mathbf{u}_{k}) \neq r_{\text{train}}( \mathbf{x}_{k+1},\mathbf{u}_{k}) \]
\end{itemize}
\textbf{Note}: In this paper, we focus on domain generalization under changes in the conditional transition probability function.
\subsection{Definition of Generalization Error in DRL}
Our goal is to quantify and analyze the generalization bound of a DRL algorithm. This is done by evaluating the performance of the trained policy under a changed environment (conditional transition probability function) compared to the training settings. The reward function is used to measure the performance of the trained DRL policy. Accordingly, we define the generalization error in DRL as: 
\begin{align}
    \text{Generalization Error} &= | \mathbb{E}_{\mathbf{u}_{k}\sim \pi, \mathbf{x}_{k+1}\sim p_{\text{test}}}[\sum^{\infty}_{k=0} \gamma_d^k r( \mathbf{x}_{k+1},\mathbf{u}_{k})] \notag \\
    &- \mathbb{E}_{\mathbf{u}_{k}\sim \pi, \mathbf{x}_{k+1}\sim p_{\text{train}}}[\sum^{\infty}_{k=0} \gamma_d^k r( \mathbf{x}_{k+1},\mathbf{u}_{k})] |,
    \label{GE}
\end{align} 
where \( \gamma_d \) is the discount factor, \( r( \mathbf{x}_{k+1},\mathbf{u}_{k}) \) is the reward function, \( \pi \) denotes the trained policy in the environment with conditional transition probability function \( p_{\text{train}} \), which corresponds to the training setting. In addition, \( p_{\text{test}} \) is the conditional transition probability function of the changed environment used for evaluation.

\subsection{Koopman Operator and DMD}
The Koopman operator theory offers a promising data-driven approach to identify and analyze the behavior of unknown nonlinear dynamical systems \cite{kutz2016dynamic}. Koopman theory was first suggested in \cite{koopman1931hamiltonian}. It demonstrates that a nonlinear dynamical system can be represented as an infinite-dimensional linear operator. 

\vspace{0.2cm}
\noindent
 \textbf{Definition 1} (\textbf{Koopman operator} \cite{kutz2016dynamic}). For a nonlinear system \( \mathbf{x}_{k+1} = f(\mathbf{x}_k) \), with \( \mathbf{x}_k \in \mathbb{R}^n \), the Koopman operator \( \mathcal{K} \) is a linear operator of infinite dimension that acts on observable functions \( g(\mathbf{x}_k)\). It satisfies the relations:
 \[\mathcal{K}g(\mathbf{x}_k) = g\circ f(\mathbf{x}_k),\]
 \[\mathcal{K}g(\mathbf{x}_k) = g(\mathbf{x}_{k+1}),\]  
 where $\circ$ denotes function composition: $g\circ f(\mathbf{x}_k) = g(f(\mathbf{x}_k)$), \( g(\mathbf{x}_k) \in \mathcal{H} \), and \( \mathcal{H} \) denotes the infinite-dimensional Hilbert space.
 
Although the Koopman operator is linear, it operates in an infinite-dimensional space, which makes it impractical for real-world applications. As a result, the applied Koopman analysis generally focuses on finite-dimensional approximations. Although various algorithms have been suggested to approximate the spectral features of Koopman operators, DMD is notably popular \cite{tu2013dynamic}. DMD estimates the Koopman operator, limited to direct observers of a system's state so that $g(\mathbf{x}_k) = \mathbf{x}_k$. Suppose the dataset driving DMD is sufficiently rich, all modes are properly excited, and the nonzero eigenvalues obtained from DMD are distinct. In that case, DMD will converge to the eigenvectors associated with the nonzero eigenvalues of the Koopman operator. Here, a sufficiently rich dataset with properly excited modes means the data captures enough time-varying behavior to represent all of the  dynamic modes of the system, allowing DMD to accurately identify the complete set of eigenvalues and modes. 
Suppose that data matrices $\mathbf{X}_0 = [\mathbf{x}_0, \mathbf{x}_1, ..., \mathbf{x}_{l-1}] \in \mathbb{R}^{n\times l}$ and $\mathbf{X}_1 = [\mathbf{x}_1, \mathbf{x}_2, ..., \mathbf{x}_{l}] \in \mathbb{R}^{n\times l}$, where the columns represent sequential snapshots of a system's state, evenly spaced in time. The procedure for the {\em standard DMD algorithm} to find DMD modes and corresponding eigenvalues of $\mathbf{\widetilde{K}}$, where $\mathbf{X}_1= \mathbf{\widetilde{K}} \mathbf{X}_0$, is \cite{ tu2013dynamic}: 
\begin{enumerate}
    \item Build a pair of data matrices \((\mathbf{X}_0,\mathbf{X}_1)\) 
    \item Compute the compact singular value decomposition (SVD) as \(\mathbf{X}_0 = \mathbf{U}_r \mathbf{S}_r \mathbf{V}_r^H\), where:
    $
    \mathbf{U}_r \in \mathbb{R}^{n \times r} \text{(left singular vectors)}, \quad \mathbf{S}_r \in \mathbb{R}^{r \times r}$  (singular values), $ \mathbf{V}_r \in \mathbb{R}^{m \times r}$(right singular vectors), and \(r = \text{rank}(\mathbf{X}_0)\) is the number of significant singular values.
    
    \item Define the reduced-order matrix \(\tilde{\mathbf{A}} = \mathbf{U}_r^H \mathbf{X}_1 \mathbf{V}_r \mathbf{S}_r^{-1}\). (This approximation represents the dynamics of $\mathbf{\widetilde{K}}$ in the reduced subspace.)
    
    \item Compute the eigenvalues \(\lambda\) and eigenvectors \(\tilde{\mathbf{v}}\) of \(\tilde{\mathbf{A}}\):
    \[
    \tilde{\mathbf{A}} \tilde{\mathbf{v}} = \lambda \tilde{\mathbf{v}}.
    \]
    \item Return the dynamic modes of $\widetilde{\mathbf{K}}$: \(\mathbf{v} = \lambda^{-1} \mathbf{X}_1 \mathbf{V}_r \mathbf{S}_r^{-1} \tilde{\mathbf{v}}\) and the corresponding eigenvalues \(\lambda\).
    \item Compute \(\mathbf{\widetilde{K}} \approx \mathbf{U}_r \widetilde{\mathbf{A}} \mathbf{U}_r^H\).
\end{enumerate}

For stochastic systems, the eigenvalues generated by the standard DMD algorithms converge to the spectrum of the Koopman operator, if the dataset driving the DMD is sufficiently rich, as long as the observables do not exhibit any randomness and are contained within a finite-dimensional invariant subspace \cite{wanner2022robust}.

The restriction on data in the DMD algorithm can be relaxed to consider data pairs
$\{(\mathbf{x}_1, \mathbf{y}_1), (\mathbf{x}_2, \mathbf{y}_2), \ldots, (\mathbf{x}_N, \mathbf{y}_N)\}$, referred to as {\em exact DMD}. Thus, the exact DMD leads to the formulation of data matrices defined as \(
\mathbf{X} = [ \mathbf{x}_1, \mathbf{x}_2, \ldots , \mathbf{x}_N]\), \( \mathbf{Y} = 
[\mathbf{y}_1, \mathbf{y}_2, \ldots , \mathbf{y}_N]\), and $\mathbf{Y}= \mathbf{\widetilde{K}} \mathbf{X}$ \cite{tu2013dynamic}. The procedure for the exact DMD algorithm is as follows:
\begin{enumerate}
\item  Arrange the data pairs into matrices $\mathbf{X}$ and $\mathbf{Y}$:
    \[
    \mathbf{X} = [\mathbf{x}_1, \mathbf{x}_2, \dots, \mathbf{x}_{m-1}], \quad \mathbf{Y} = [\mathbf{y}_1, \mathbf{y}_2, \dots, \mathbf{y}_{m-1}].
    \]
\item  Compute the reduced SVD of $\mathbf{X}$:
    $\mathbf{X} = \mathbf{U} \mathbf{\Sigma} \mathbf{V}^H.$
\item  Define the matrix $\tilde{\mathbf{A}}$:
$ \tilde{\mathbf{A}} = \mathbf{U}^H \mathbf{Y} \mathbf{V} \mathbf{\Sigma}^{-1}.$
\item  Compute the eigenvalues and eigenvectors of $\tilde{\mathbf{A}}$:
    \[
    \tilde{\mathbf{A}} \mathbf{v} = \lambda \mathbf{v}.
    \]
    Each nonzero eigenvalue $\lambda$ is a DMD eigenvalue.
\item  The DMD mode corresponding to $\lambda$ is then obtained as:
    \[
    \boldsymbol{\phi} = \frac{1}{\lambda} \mathbf{Y} \mathbf{V} \mathbf{\Sigma}^{-1} \mathbf{v}.
    \]
\end{enumerate}
\noindent
\textbf{Theorem 1} \cite{tu2013dynamic}. Each pair $(\boldsymbol{\phi},\lambda)$ produced by the exact DMD algorithm is an eigenvector/eigenvalue pair of \(\widetilde{\mathbf{K}}\). Furthermore, the algorithm identifies all the nonzero eigenvalues of \(\mathbf{\widetilde{K}}\).
\subsection{$Z$-Transformation and H$_{\infty}$ Norm}
The \( Z \)-transform technique is a mathematical tool widely used in scientific and engineering fields for analyzing and understanding the dynamic behavior of discrete-time systems. It transforms the difference equations in the time domain into algebraic equations in the frequency domain, simplifying the system analysis. By converting the system equations into the \( Z \)-domain, we can study the overall dynamic behavior of discrete-time systems under various input conditions. The \( Z \)-transform of a discrete causal signal, \( \mathbf{x}_k \), defined for all integer values of \( k \), \( k \geq 0 \), is given by \cite{oppenheim1999signals}:
\begin{equation}
    Z\{\mathbf{x}_k\} = \mathbf{x}_z = \sum_{k=0}^{\infty} \mathbf{x}_k z^{-k}.
\end{equation}
The H$_{\infty}$ norm is a well-established metric in control theory for quantifying a system’s worst-case gain across all frequencies. 
Specifically, the H$_{\infty}$ norm represents the maximum possible magnitude of a transfer function across all frequencies. It corresponds to the system’s worst-case response to an input. For a system with a transfer function \(\mathbf{K}_z\), the H\(_\infty \) norm is given by \cite{optimalrobust}:
\begin{equation}
\|\mathbf{K}_z \|_{H_\infty} = \sup_{\omega \in [0, \pi]} \sigma_{\text{max}}(\mathbf{K}_z(e^{j\omega})),
\end{equation}
where \(\mathbf{K}_z(e^{j\omega}) \) is the transfer function evaluated on the unit circle \( z = e^{j\omega} \), \( \sigma_{\text{max}}(\mathbf{K}_z(e^{j\omega})) \) is the maximum singular value of \(\mathbf{K}_z(e^{j\omega}) \), and \( \omega \) represents the normalized frequency (ranging from 0 to \( \pi \)). The singular values of a matrix \(\mathbf{K}_z\) are defined as the square roots of the eigenvalues of \(\mathbf{K}_z^H\mathbf{K}_z\).
\section{Identifying Dynamic Behavior of Deep Reinforcement Learning}
In this section, we first model the evolution of states and actions in a trained DRL algorithm as unknown discrete stochastic nonlinear dynamics. We then introduce an additive disturbance vector to represent domain changes. Finally, we use the Koopman operator and DMD to identify the unknown dynamics.
\subsection{Dynamical System Model of Deep Reinforcement Learning}
\begin{figure}[t!] 
    \centering
    \includegraphics[width=3.4 in]{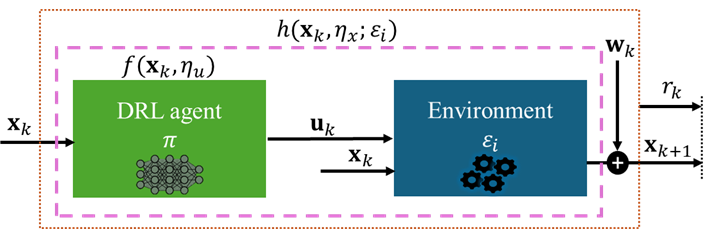}
    \caption{Dynamical system model of deep reinforcement learning}
    \label{DRL}
\end{figure}

A DRL involves an agent interacting with environment $\varepsilon_i \in \mathcal{S}$, transitioning through a series of states $\mathbf{x}_k \in \mathbb{R}^n$, and taking actions $\mathbf{u}_k \in \mathbb{R}^m$ at each time step $k \in \mathscr{K} = \{0,1,...,K-1\}$. In the trained DRL, the action is sampled from a trained offline policy $\mathbf{u}_k \sim \pi$ and executed in environment $\varepsilon_i$. As shown in Fig.~\ref{DRL}, this action leads to a new state $\mathbf{x}_{k+1}$ and generates a reward $r_k = r(\mathbf{x}_{k+1},\mathbf{u}_k) \in \mathbb{R}$, where $r$ is a predefined known function. 
Despite the black-box nature of $\pi$ and the unknown conditional transition probability function of $\varepsilon_i$, it is possible to represent the evolution associated with $\mathbf{x}_k$ and $\mathbf{u}_k$ as discrete dynamical stochastic nonlinear systems:
\begin{equation}
     \mathbf{u}_k = f (\mathbf{x}_{k},\mathbf{\eta}_u),\\
    \label{sys1}
\end{equation}
\begin{equation}
    \mathbf{x}_{k+1} = h(\mathbf{x}_{k},\mathbf{\eta}_x; \varepsilon_i), 
    \label{sys3}
\end{equation}
\noindent where $f$ and $h$ are unknown nonlinear functions, $\mathbf{\eta}_u$ and $\mathbf{\eta}_x$ are random variables that introduce randomness into dynamical systems, and $\varepsilon_i$ represents different environments $\varepsilon_i \in \mathcal{S}$. 
Indeed, $\mathcal{S}$ denotes the space of all possible environments. These representations capture the dynamics of the decision-making policy (green box in Fig.~\ref{DRL}) and the state (pink dotted box in Fig.~\ref{DRL}) of the DRL agent.

In addition, it is important to mention that DRL algorithms can be grouped into two categories: value-based and policy-based methods. The policy-based methods can be further divided into stochastic and deterministic policies. Additionally, value-based methods are considered deterministic policies. Therefore, in DRL, a distinction is made between stochastic and deterministic policies, and when the policy is deterministic, $\mathbf{\eta}_u$ is equal to zero.

\vspace{0.2cm}
\noindent
\textbf{Assumption 1}. Each environment  $\varepsilon_i \in \mathcal{S}$ has a unique and unknown conditional transition probability function, and $p_i$ represents the conditional transition probability function of environment $\varepsilon_i$. 

\vspace{0.2cm}
\noindent
\textbf{Assumption 2}. Given any DRL policy $\pi$, $r(\mathbf{x}_{k+1}, \mathbf{u}_k)$ is known and fixed. 
\subsection{Modeling Domain Changes Using Additive Disturbance}
We model domain changes over the conditional transition probability function of an environment ($\mathbf{x}_{k+1}\sim p_i\{\mathbf{x}_{k+1}|\mathbf{x}_{k},\mathbf{u}_{k}\}$) by an additive disturbance vector:  
\begin{align}
    \mathbf{x}^{\text{w}}_{k+1} &= \mathbf{x}_{k+1} + \mathbf{w}_k,
    \label{domain_model}
\end{align}
where $\mathbf{w}_k\sim p_{w}$ is a random disturbance vector. Therefore, $\mathbf{x}^{\text{w}}_{k+1}\sim p^{\text{w}}_i\{\mathbf{x}^{\text{w}}_{k+1}|\mathbf{x}_{k},\mathbf{u}_{k}\}$, where $p^{\text{w}}_i = p_i \circledast p_{w}$, and $\circledast $ denotes convolution operator.

\vspace{0.2cm}
\noindent
\textbf{Assumption 3}. We assume that $\mathbf{w}_k$ and $\mathbf{x}_{k+1}$ are independent random variables. The random disturbance vector $\mathbf{w}_k$ follows an unknown distribution $p_{w}$, i.e., $\mathbf{w}_k \sim p_{w}$. 

Accordingly, the stochastic nonlinear model associated with the state evolution in (\ref{sys3}) is modified as:
\begin{align}
    \mathbf{x}^{\text{w}}_{k+1} &= h(\mathbf{x}_{k},\mathbf{\eta}_x) + \mathbf{w}_k, 
\end{align}
\subsection{Using Koopman Operator and DMD to Identify Unknown Dynamical Functions}

In Section III.A, we modeled evolution associated with $\mathbf{x}_k$ and $\mathbf{u}_k$ as the discrete stochastic nonlinear dynamics (\ref{sys1}) and (\ref{sys3}). However, the nonlinear dynamics are unknown. 
Here, we first apply Koopman operator theory, which operates on observable functions of the dynamics' states in equations (\ref{sys1}) and (\ref{sys3}). This data-driven approach enables us to identify and analyze the unknown nonlinear dynamics from a linear viewpoint. Then, we use DMD and exact DMD to approximate the Koopman operators.

Assume observer functions $g(\mathbf{x}_k)$ and $g(\mathbf{u}_k)$ for both $\mathbf{x}_k$ and $\mathbf{u}_k$, the Koopman operators for systems (\ref{sys1}) and (\ref{sys3}) are: 
\begin{align}
g(\mathbf{u}_{k}) &= \mathcal{K}^f g(\mathbf{x}_k),\\
g(\mathbf{x}_{k+1}) &= \mathcal{K}^h g(\mathbf{x}_{k}),
\end{align}
where $\mathcal{K}^f$ and $\mathcal{K}^h$ are the Koopman operators for systems (\ref{sys1}) and (\ref{sys3}), respectively. Now, we employ the exact DMD and DMD to approximate the spectral features of $\mathcal{K}^f$ and $\mathcal{K}^h$. Accordingly, observer function $g$ for both $\mathbf{x}_k$ and $\mathbf{u}_k$ is considered as the expected value of the variables over multiple independent realizations of the trained DRL:
\begin{align}
    g(\mathbf{x}_k) &= \overline{\mathbf{x}}_{k}, \\
    g(\mathbf{u}_k) &= \overline{\mathbf{u}}_{k},
\end{align}
where $\overline{\mathbf{x}}_{k}$ and $\overline{\mathbf{u}}_{k}$ respectively represent the expected values of $\mathbf{x}_{k}$ and $\mathbf{u}_{k}$ over multiple independent realizations, defined as $\mathbb{E}_{\pi,p_i} [\mathbf{x}_k]$ and $\mathbb{E}_{\pi,p_i} [\mathbf{u}_k]$. 
Thus, we can approximate the expected evolution of $\mathbf{u}_k$ and $\mathbf{x}_k$ as:
\begin{align}
    \overline{\mathbf{u}}_k &= \mathbf{\widetilde{K}}^f \overline{\mathbf{x}}_{k} ,\label{inter_u}  \\
    \overline{\mathbf{x}}_{k+1} &=\mathbf{\widetilde{K}}^h \overline{\mathbf{x}}_{k} ,\label{inter_general state}
\end{align}
where $\mathbf{\widetilde{K}}^f $ and $\mathbf{\widetilde{K}}^h$ represent approximated $\mathcal{K}^f$ and $\mathcal{K}^h$ using the exact DMD and DMD, respectively. It is worth emphasizing that DMD eigenvalues converge to the Koopman spectrum for stochastic systems if the dataset is rich and the observables remain free of randomness \cite{wanner2022robust}. Accordingly, we consider $g(\mathbf{x}_k) = \overline{\mathbf{x}}_k$ and $g(\mathbf{u}_k) = \overline{\mathbf{u}}_k$ as observer functions for $\mathbf{x}_k$ and $\mathbf{u}_k$, respectively.

Equations (\ref{inter_u}) and (\ref{inter_general state}) provide interpretable representations of the DRL dynamics based on the expected values of the DRL's variables. Recall that, in Section III.B, we modeled the domain changes as an additive disturbance. To incorporate the domain changes, the interpretable DRL model (\ref{inter_general state}) is adjusted as follows:
\begin{align}
    \overline{\mathbf{x}}^{\text{w}}_{k+1} &=\mathbf{\widetilde{K}}^h \overline{\mathbf{x}}_{k} + \overline{\mathbf{w}}_k,\label{inter_general state_n}
\end{align}
where $\bar{\mathbf{w}}_k= \mathbb{E}_{p_w} [\mathbf{w}_k]$ is the expected value of ${\mathbf{w}}_k$ at time $k$.
Fig.~2 shows a visual illustration of the proposed interpretable models for DRLs.
\begin{figure}[t] 
    \centering
    \includegraphics[width=3.4 in]{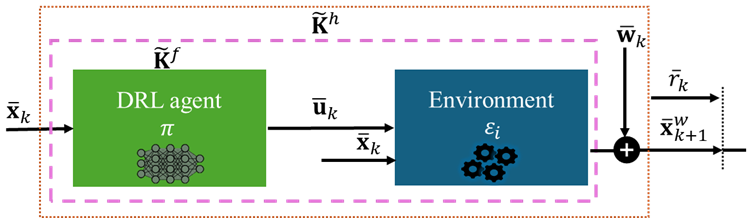}
    \caption{Interpretable model for deep reinforcement learning}
    \label{DRL_interpretable}
\end{figure}
\section{Method for Generalizability Analysis in DRL}
In Section III, we have presented interpretable models for the evolution of state and action in DRL. In this section, we propose a method for quantifying the generalizability bound of a trained DRL policy using those interpretable models. Specifically, we analyze the generalization bound by evaluating the performance of the trained policy under a changed conditional transition probability function compared to the training settings.  
\subsection{Impact of Domain Changes on System Dynamics}
In this subsection, we estimate how a domain change can impact a trained DRL's states and actions in the worst-case scenario. Specifically, the H$_\infty$ norm is used to evaluate the DRL's robustness to domain changes.

First, to analyze the dynamic behavior of the DRL under the distribution changes of the environment, we transfer the interpretable models (\ref{inter_u}) and (\ref{inter_general state_n}) into the \( Z \)-domain:
\begin{align}
    \overline{\mathbf{u}}_z &= \mathbf{\widetilde{K}}^f \overline{\mathbf{x}}_z, \label{Z1} \\
    z\overline{\mathbf{x}}^{\text{w}}_z -z\overline{\mathbf{x}}^{\text{w}}_{k=0} &= \mathbf{\widetilde{K}}^h \overline{\mathbf{x}}_z + \overline{\mathbf{w}}_z. \label{Z3}
\end{align}
Accordingly, the transfer function from $\overline{\mathbf{w}}_z$ and $\overline{\mathbf{x}}^{\text{w}}_{k=0}$ to $\overline{\mathbf{x}}^{\text{w}}_z$ can be calculated as:
\begin{align*}
    z\overline{\mathbf{x}}^{\text{w}}_z -\mathbf{\widetilde{K}}^h \overline{\mathbf{x}}^{\text{w}}_z =  z\overline{\mathbf{x}}^{\text{w}}_{k=0}+ \overline{\mathbf{w}}_z,
\end{align*}
\begin{align*}
    (z\mathbf{I}-\mathbf{\widetilde{K}}^h)\overline{\mathbf{x}}^{\text{w}}_z =  z\overline{\mathbf{x}}^{\text{w}}_{k=0}+ \overline{\mathbf{w}}_z,
\end{align*}
\begin{equation}
    \overline{\mathbf{x}}^{\text{w}}_z =  (z\mathbf{I}-\mathbf{\widetilde{K}}^h)^{-1}(z\overline{\mathbf{x}}^{\text{w}}_{k=0})  +  {(z\mathbf{I}-\mathbf{\widetilde{K}}^h)}^{-1}\overline{\mathbf{w}}_z. \label{ZT}
\end{equation}
Hereafter, we denote the expected value of the DRL's state without/with the domain change by $\overline{\mathbf{x}}^n$ and $\overline{\mathbf{x}}^{\text{w}}$, respectively. \\
\vspace{-5pt}

\noindent\textbf{Assumption 4}: We assume that $\overline{\mathbf{x}}^n_{k=0} = \overline{\mathbf{x}}^{\text{w}}_{k=0}$.\\
\vspace{-5pt}

By considering equation (\ref{ZT}) and \textbf{Assumption 4}, we have: 
\begin{equation}
\overline{\mathbf{x}}_z^n - \overline{\mathbf{x}}_z^{\text{w}} = {(z\mathbf{I} - \mathbf{\widetilde{K}}^h)}^{-1}\overline{\mathbf{w}}_z. 
\label{transfer_1}
\end{equation}
Therefore, the transfer function matrix from $\overline{\mathbf{w}}_z$ to  \(\overline{\mathbf{x}}_z^n - \overline{\mathbf{x}}_z^{\text{w}}\) is: 
\begin{equation}
{\mathbf{T}}_z^{{\text{w}}n} = {(z\mathbf{I} - \mathbf{\widetilde{K}}^h)}^{-1}.
\label{transfer_2}
\end{equation}
Accordingly, the H\( _\infty \) norm of the transfer function \({\mathbf{T}}_z^{{\text{w}}n} \) is:
\begin{equation}
\| {\mathbf{T}}_z^{\text{w}n}  \|_{\text{H}_\infty} = \sup_{\omega \in [0, \pi]} \sigma_{\text{max}}\left( {(e^{j\omega}\mathbf{I} - \mathbf{\widetilde{K}}^h)^{-1}} \right).
\end{equation}

\vspace{2pt}
\noindent
\textbf{Theorem 2.} Given a trained DRL policy, for any domain change such that \( \|{\mathbf{w}}_z\|_{H_\infty} \leq \gamma \),  the term \( \|\mathbf{T}^{\text{w}n}_z\|_{{H}_\infty} \) directly influence the maximum impact that such domain changes can have on the DRL policy's states:
\begin{equation}
\sum_{k=0}^{K-1} (\|\overline{\mathbf{x}}_k^n - \overline{\mathbf{x}}_k^{\text{w}}\|_2)^2 \leq (\|{\mathbf{T}}_z^{{\text{w}}n}\|_{H_\infty} \cdot \gamma)^2, 
\end{equation}
\begin{equation}
\max_{k \in \mathscr{K}} \|\overline{\mathbf{x}}_k^n - \overline{\mathbf{x}}_k^{\text{w}}\|_2 \leq \|{\mathbf{T}}_z^{\text{w}n}\|_{H_\infty} \cdot \gamma.
\end{equation}

\vspace{0.2cm}
\noindent
\textbf{Proof}.
The given condition is:
\[
\|{\mathbf{w}}_z\|_{H_\infty} \leq \gamma,
\]
indicating that:
\[
\sup_{\omega \in [0, 2\pi]} \sigma_{\text{max}}({\mathbf{w}}_z(e^{j\omega})) \leq \gamma,
\]
where \( {\mathbf{w}}_z \) is a vector. Treating \( {\mathbf{w}}_z \) as a matrix of size \(n \times 1\), the singular values of \( {\mathbf{w}}_z \) are the square roots of the eigenvalues of \({\mathbf{w}}_z^T {\mathbf{w}}_z\). Compute \({\mathbf{w}}_z^T {\mathbf{w}}_z\) as:
\[
{\mathbf{w}}_z^T {\mathbf{w}}_z = \| {\mathbf{w}}_z\|_2^2.
\]
The only singular value of \({\mathbf{w}}_z\) is therefore:
\[
\sigma_{\text{max}}({\mathbf{w}}_z) = \sqrt{\| {\mathbf{w}}_z\|_2^2} = \|{\mathbf{w}}_z \|_2.
\]
Thus, we have:
\[
\sup_{\omega \in [0, 2\pi]} \|{\mathbf{w}}_z(e^{j\omega})\|_2 \leq \gamma.
\]
Then, using Jensen's inequality and the convexity of the norm:
\begin{align*}
 \| \bar{\mathbf{w}}_z(e^{j\omega}) \|_2 = \left\| \mathbb{E}[ \mathbf{w}_z(e^{j\omega}) ] \right\|_2 
&\leq \mathbb{E} \left[ \| \mathbf{w}_z(e^{j\omega}) \|_2 \right] \\
&\leq \sup_{\omega} \| \mathbf{w}_z(e^{j\omega}) \|_2 = \| \mathbf{w}_z \|_{\mathcal{H}_\infty}.   
\end{align*}
\noindent Therefore, we conclude:
\[
\| \bar{\mathbf{w}}_z \|_{\mathcal{H}_\infty} \leq \gamma.
\]
By considering equations (\ref{transfer_1}) and (\ref{transfer_2}), we have:
\[
\overline{\mathbf{x}}_z^n- \overline{\mathbf{x}}_z^{\text{w}}= {\mathbf{T}}_z^{\text{w}n} \overline{\mathbf{w}}_z,
\]
where \(\overline{\mathbf{x}}_z^n\), \(\overline{\mathbf{x}}_z^{\text{w}}\), and  \( \overline{\mathbf{w}}_z\) are vectors in the $Z$-domain and \({\mathbf{T}}_z^{\text{w}n}\) is a matrix in the $Z$-domain. 
We aim to calculate \( \|\overline{\mathbf{x}}_z^n- \overline{\mathbf{x}}_z^{\text{w}} \|_{H_\infty} \), which is given by:
\[
\|\overline{\mathbf{x}}_z^n- \overline{\mathbf{x}}_z^{\text{w}} \|_{H_\infty} = \|{\mathbf{T}}_z^{\text{w}n} \overline{\mathbf{w}}_z\|_{H_\infty}.
\]
Using the sub-multiplicative property of $H_\infty$ norms, we can state:
\[
\|{\mathbf{T}}_z^{\text{w}n} \overline{\mathbf{w}}_z\|_{H_\infty} \leq \|{\mathbf{T}}_z^{\text{w}n}\|_{H_\infty} \| \overline{\mathbf{w}}_z\|_{H_\infty}.
\]
Since \( \|\overline{\mathbf{w}}_z\|_{H_\infty} \leq \gamma \), its maximum possible impact on \( \bar{\mathbf{x}}_z^n - \overline{\mathbf{x}}_z^{\text{w}} \) is:
\[
\|\overline{\mathbf{x}}_z^n- \overline{\mathbf{x}}_z^{\text{w}} \|_{H_\infty} \leq \|{\mathbf{T}}_z^{\text{w}n}\|_{H_\infty} \cdot \gamma.
\]
As $\overline{\mathbf{x}}_z^n - \overline{\mathbf{x}}_z^{\text{w}}$ is a vector, we can apply an analysis similar to that used for $\overline{\mathbf{w}}_z$  mentioned above, yielding:
\[
\sup_{\omega \in [0, 2\pi]} \|\overline{\mathbf{x}}_z^n(e^{j\omega}) - \overline{\mathbf{x}}_z^{\text{w}}(e^{j\omega})\|_2 \leq \|{\mathbf{T}}_z^{\text{w}n}\|_{H_\infty} \cdot \gamma.
\]
Now, to represent the above bound in the time domain, Parseval's theorem is used:
\[
\sum_{k=0}^{K-1} \|\overline{\mathbf{x}}_k^n- \overline{\mathbf{x}}_k^{\text{w}}\|_2^2 = \frac{1}{2\pi} \int_{0}^{2\pi} \|\overline{\mathbf{x}}_z^n(e^{j\omega}) - \overline{\mathbf{x}}_z^{\text{w}}(e^{j\omega})\|_2^2 \; d\omega.
\]
By considering \(\sup_{\omega \in [0, 2\pi]} \|\overline{\mathbf{x}}_z^n(e^{j\omega}) - \overline{\mathbf{x}}_z^{\text{w}}(e^{j\omega})\|_2 \leq \|\mathbf{T}^{\text{w}n}_z\|_{H_\infty} \cdot \gamma \) and Parseval's theorem, we have: 
\[
\sum_{k=0}^{K-1} \|\overline{\mathbf{x}}_k^n- \overline{\mathbf{x}}_k^{\text{w}}\|_2^2 \leq (\|{\mathbf{T}}_z^{\text{w}n}\|_{H_\infty} \cdot \gamma)^2.
\]
Furthermore, since each term in the summation $\sum_{k=0}^{K-1} \|\overline{\mathbf{x}}_k^n- \overline{\mathbf{x}}_k^{\text{w}}\|_2^2$ is non-negative, we have:
\[
 \max_{k \in \mathscr{K}}(\|\overline{\mathbf{x}}^n_k-\overline{\mathbf{x}}^{\text{w}}_k\|_2 ) \leq \|{\mathbf{T}}_z^{\text{w}n}\|_{H_\infty} \cdot \gamma. \qed
\]

\noindent
\textbf{Interpretation of $\| {\mathbf{w}}_z\|_{H_\infty} \leq \gamma$ in time domain}: Using Parseval's theorem, we relate the characteristic of domain change ${\mathbf{w}}_z$ in the time domain:
\[
\sum_{k=0}^{K-1} \|{\mathbf{w}}_k\|_2^2 = \frac{1}{2\pi} \int_{0}^{2\pi} \| {\mathbf{w}}_z(e^{j\omega})\|_2^2 \; d\omega.
\]
Given \( \|{\mathbf{w}}_z\|_{H_\infty} \leq \gamma \), we have
\( \sup_{\omega \in [0, 2\pi]} \|{\mathbf{w}}_z(e^{j\omega})\|_2 \leq \gamma \), so:
\[
\sum_{k=0}^{K-1} \|{\mathbf{w}}_k\|_2^2 \leq \frac{1}{2\pi} \int_{0}^{2\pi} \gamma^2 d\omega = \gamma^2.
\]
It can be interpreted that $ \gamma^2$ is a bound on the total energy of the ${\mathbf{w}}_k$ over time. Moreover, we can derive that:
\begin{equation}
    \|{\mathbf{w}}_k\|_2 \leq  \gamma, \quad \forall k \in \mathscr{K}.
    \label{ex_dis}
\end{equation}
This means that the Euclidean norm of $\mathbf{w}_k$ must satisfy equation~(\ref{ex_dis}) at all times.

\noindent \textbf{Interpretation of $\| {\mathbf{w}}_z \|_{H_\infty} \leq \gamma$ using Wasserstein distance}:
Reconsider equation (\ref{domain_model}):
\[
\mathbf{x}_{k+1}^{\text{w}} = \mathbf{x}_{k+1} + \mathbf{w}_k, \quad \mathbf{w}_k \sim p_w,
\]
which results in the domain change in the conditional transition probability function of the environment:
\[
p_i^{\text{w}}(\mathbf{x}_{k+1}^{\text{w}} \mid \mathbf{x}_k, \mathbf{u}_k) = p_i(\cdot \mid \mathbf{x}_k, \mathbf{u}_k) \circledast p_w.
\]
Using the triangle inequality for Wasserstein-1 distance, we get:
\[
W_1(p_i^{\text{w}}, p_i) = W_1(p_i \circledast p_w, p_i) \leq W_1(p_w, \delta_0),
\]
where $\delta_0$ is the Dirac delta at the origin. This means that the change in the conditional transition probability function of an environment due to domain change is at most the Wasserstein distance between $p_w$ and the origin.

\noindent Using the interpretation of $\| {\mathbf{w}}_z \|_{H_\infty} \leq \gamma$ in time domain:
\[
\|{\mathbf{w}}_k \|_2 \leq \gamma, \quad \forall k \in \mathscr{K},
\]
and since $W_1(p_w, \delta_0) \leq \mathbb{E}[\|\mathbf{w}_k\|_2] \leq \gamma$, we can conclude:
\[
W_1(p_i^{\text{w}}, p_i) \leq \gamma.
\]
In other words, the bounding $\|{\mathbf{w}}_z \|_{H_\infty}$ directly controls the Wasserstein distance between the changed and nominal conditional transition probability function of the environment.

%
\vspace{0.2cm}
\noindent
\textbf{Corollary 1}. Given a trained DRL policy, for any domain change such that \( \|{\mathbf{w}}_z\|_{{H}_\infty} \leq \gamma \), the terms \( \|\mathbf{T}^{\text{w}n}_z\|_{{H}_\infty} \) and \( \| \widetilde{\mathbf{K}}^f_z\|_{{H}_\infty} \) directly influence the maximum impact that such domain changes can have on the DRL policy's actions: 
\begin{equation}
\sum_{k=0}^{K-1} \|\overline{\mathbf{u}}_k^n- \bar{\mathbf{u}}_k^{\text{w}}\|_2^2  \leq (\| {\mathbf{\widetilde{K}}^f}_z \|_{H_\infty} \cdot \|\mathbf{T}^{\text{w}n}_z\|_{H_\infty} \cdot \gamma)^2.
\label{energy_u}
\end{equation}
\begin{equation}
\max_{k \in \mathscr{K}}(\|\overline{\mathbf{u}}_k^n- \overline{\mathbf{u}}_k^{\text{w}}\|_2) \leq \| {\mathbf{\widetilde{K}}^f}_z \|_{H_\infty} \cdot \|\mathbf{T}^{\text{w}n}_z\|_{H_\infty} \cdot \gamma .  
\end{equation}

\vspace{0.2cm}
\noindent
\textbf{Proof}.
H\(_\infty \) norm of $\mathbf{\widetilde{K}}^f$ is defined as $\| {\mathbf{\widetilde{K}}^f}_z \|_{H_\infty} = \sup_{\omega \in [0, \pi]} \sigma_{\text{max}}({\mathbf{\widetilde{K}}^f}_z (e^{j\omega}))$.
Therefore, by considering equation (\ref{inter_u}), (\ref{transfer_1}), and the sub-multiplicative property of $H_\infty$ norms, we have:
\begin{equation}
\|\overline{\mathbf{u}}_z^n- \overline{\mathbf{u}}_z^{\text{w}}\|_{H_\infty} \leq \| {\mathbf{\widetilde{K}}^f}_z \|_{H_\infty} \cdot \|\mathbf{T}^{\text{w}n}_z\|_{H_\infty} \cdot \gamma. 
\end{equation}
Similarly, Parseval’s theorem can be used to relate the characteristics of the signal $\|\overline{\mathbf{u}}_z^n- \overline{\mathbf{u}}_z^{\text{w}}\|_{H_\infty}$ in the frequency domain to its representation in the time domain:
\begin{equation*}
\sum_{k=0}^{K-1} \|\overline{\mathbf{u}}_k^n- \overline{\mathbf{u}}_k^{\text{w}}\|_2^2  \leq (\| {\mathbf{\widetilde{K}}^f}_z \|_{H_\infty} \cdot \|\mathbf{T}^{\text{w}n}_z\|_{H_\infty} \cdot \gamma)^2.
\end{equation*}
Equation (\ref{energy_u}) provides an energy constraint on the maximum effect of domain changes on the DRL policy's action. 
Moreover, each term in the summation $\sum_{k=0}^{K-1} \|\overline{\mathbf{u}}_k^n- \overline{\mathbf{u}}_k^{\text{w}}\|_2^2$ is non-negative, therefore:
\[\max_{k \in \mathscr{K}}(\|\overline{\mathbf{u}}_k^n- \overline{\mathbf{u}}_k^{\text{w}}\|_2) \leq \| {\mathbf{\widetilde{K}}^f}_z \|_{H_\infty} \cdot \|\mathbf{T}^{\text{w}n}_z\|_{H_\infty} \cdot \gamma .\qed\] 
\vspace{-10pt}

\subsection{Analysis of Generalizability}
In this subsection, we aim to evaluate the
generalizability of the trained DRL policy's performance. We estimate the maximum effect of domain changes on the DRL performance. Specifically, we analyze the performance of DRL in terms of reward function. Based on \textbf{Assumption 2}, the reward function is assumed to be known and fixed and expressed as \( r(\mathbf{x}_{k+1}, \mathbf{u}_k) \), a function of \( \mathbf{x}_{k+1} \) and \( \mathbf{u}_k \). In \textbf{Theorem 2} and \textbf{Corollary 1}, we estimated the maximum impact of domain changes on the trained DRL policy's state and action variables. Therefore, using the known relationship between the state, action, and reward, we can derive the maximum impact of domain changes on the reward function. Moreover, we calculate a maximum bound on the generalization error of a trained DRL. 
\vspace{2pt}

\noindent
\textbf{Assumption 5}: The reward function of the DRL satisfies the Lipschitz condition with Lipschitz constant $L$.

\vspace{0.2cm}
\noindent
\textbf{Definition 2}: A function \( f(\mathbf{x}, \mathbf{u}) \) satisfies a \textbf{Lipschitz condition} if there exists a constant \( L \) such that:
\[
|f(\mathbf{x}_1, \mathbf{u}_1) - f(\mathbf{x}_2, \mathbf{u}_2)| \leq L \left( \|\mathbf{x}_1 - \mathbf{x}_2\|_2 + \|\mathbf{u}_1 - \mathbf{u}_2\|_2 \right),
\]
for all pairs of inputs \( (\mathbf{x}_1, \mathbf{u}_1) \) and \( (\mathbf{x}_2, \mathbf{u}_2) \) within the domain of \( f \). Here, \( L \) is called the \textbf{Lipschitz constant}, which essentially bounds the rate of change of \( f \) with respect to changes in \( \mathbf{x} \) and \( \mathbf{u} \).

\vspace{5pt}
\noindent
\textbf{Corollary 2}. Given a trained DRL policy, for any domain change satisfying $\|{\mathbf{w}}_z\|_{H_\infty} \leq \gamma$, the terms $\|\mathbf{T}^{\text{w}n}_z\|_{H_\infty} $ and $\| {\mathbf{\widetilde{K}}^f}_z\|_{H_\infty}$ directly influence the magnitude of the maximum impact that domain changes have on the expected cumulative reward of the trained policy.\\ 

\noindent
\textbf{Proof}.
According to \textbf{Assumption 5}, $r(\mathbf{x}_{k+1}, \mathbf{u}_k)$ satisfies the Lipschitz condition with Lipschitz constant $L$. Therefore, we have:
\begin{align}
    |r(\mathbf{x}^{\text{w}}_{k+1}, \mathbf{u}^{\text{w}}_k) - r(\overline{\mathbf{x}}^{\text{w}}_{k+1}, \overline{\mathbf{u}}^{\text{w}}_k)| \leq &L (\|\mathbf{x}^{\text{w}}_{k+1} - \overline{\mathbf{x}}^{\text{w}}_{k+1}\|_2 \nonumber\\ &+ \|\mathbf{u}^{\text{w}}_k - \bar{\mathbf{u}}^{\text{w}}_k\|_2),
    \label{l1}
\end{align}
and
\begin{equation}
    |r(\overline{\mathbf{x}}^{\text{w}}_{k+1}, \overline{\mathbf{u}}^{\text{w}}_k) - r(\overline{\mathbf{x}}^n_k, \overline{\mathbf{u}}^n_k)| \leq L (\|\overline{\mathbf{x}}^{\text{w}}_{k+1} - \overline{\mathbf{x}}^n_{k+1}\|_2 + \|\overline{\mathbf{u}}^{\text{w}}_k - \overline{\mathbf{u}}^n_k\|_2).
    \label{l2}
\end{equation}

\noindent Let $M=\|\mathbf{T}^{\text{w}n}_z\|_{H_\infty} \cdot \gamma$ and $N=\| {\mathbf{\widetilde{K}}^f}_z\|_{H_\infty} \cdot \|\mathbf{T}^{\text{w}n}_z\|_{H_\infty} \cdot \gamma$. Considering equations (\ref{l1}), (\ref{l2}), \textbf{Theorem 2} and \textbf{Corollary 1}, and the triangle inequality, we have:
\begin{align}
|r(\mathbf{x}^{\text{w}}_{k+1}, \mathbf{u}^{\text{w}}_k) - r(\overline{\mathbf{x}}^n_{k+1}, \overline{\mathbf{u}}^n_k)| \leq& L ((\|\mathbf{x}^{\text{w}}_{k+1} - \overline{\mathbf{x}}^{\text{w}}_{k+1}\|_2 \nonumber\\+& \|\mathbf{u}^{\text{w}}_k - \overline{\mathbf{u}}^{\text{w}}_k\|_2)+(M+N)),
\label{l3}
\end{align}
Taking expectations on both sides:
\begin{align}
    \mathbb{E}_{\pi,p^{\text{w}}}[|r(\mathbf{x}^{\text{w}}_{k+1}, \mathbf{u}^{\text{w}}_k)-&r(\overline{\mathbf{x}}^n_{k+1}, \overline{\mathbf{u}}^n_k)|] \leq L( \nonumber\\ \mathbb{E}_{\pi,p^{\text{w}}} [(\|\mathbf{x}^{\text{w}}_{k+1} - \overline{\mathbf{x}}^{\text{w}}_{k+1}\|_2+& \|\mathbf{u}^{\text{w}}_k - \overline{\mathbf{u}}^{\text{w}}_k\|_2)+(M+N)]).
\end{align}
Let $\mathbb{E}_{\pi,p^{\text{w}}} [\|\mathbf{x}^{\text{w}}_{k+1} - \overline{\mathbf{x}}^{\text{w}}_{k+1}\|_2 + \|\mathbf{u}^{\text{w}}_{k} - \overline{\mathbf{u}}^{\text{w}}_{k}\|_2]=Q$. The absolute value function is convex, and by applying Jensen's inequality, we obtain:
\begin{align}
    |\mathbb{E}_{\pi,p^{\text{w}}}[r(\mathbf{x}^{\text{w}}_{k+1}, \mathbf{u}^{\text{w}}_k) - r(\overline{\mathbf{x}}^n_{k+1}, \overline{\mathbf{u}}^n_k)]|& \leq L(Q \nonumber\\& +(M+N)),\nonumber\\
    |\mathbb{E}_{\pi,p^{\text{w}}}[r(\mathbf{x}^{\text{w}}_{k+1}, \mathbf{u}^{\text{w}}_k)] - r(\overline{\mathbf{x}}^n_{k+1}, \overline{\mathbf{u}}^n_k)|& \leq L(Q \nonumber\\& +(M+N)).
\end{align}
Now, summing over all time steps yields:
\begin{equation*}
    \sum^{K-1}_{k=0} |\mathbb{E}_{\pi,p^{\text{w}}}(r(\mathbf{x}^{\text{w}}_{k+1}, \mathbf{u}^{\text{w}}_k)) - r(\bar{\mathbf{x}}^n_{k+1}, \overline{\mathbf{u}}^n_k)|\leq
\end{equation*}
\begin{equation*}
    \sum^{K-1}_{k=0} L (Q+M + N) = L (Q+M + N) \sum^{K-1}_{k=0} 1.
\end{equation*}
For a discount factor \( \gamma_d \), we can write:
\begin{equation*}
    \sum^{K-1}_{k=0} \gamma_d^k = \frac{1- \gamma_d^{K+1}}{1 - \gamma_d},
\end{equation*}
and
\begin{equation*}
    \sum^{\infty}_{k=0} \gamma_d^k = \frac{1}{1 - \gamma_d}, \quad  0 \leq \gamma_d < 1.
\end{equation*}
Thus:
\begin{align*}
    \sum^{K-1}_{k=0} \gamma_d^k |\mathbb{E}_{\pi,p^{\text{w}}}[r(\mathbf{x}^{\text{w}}_{k+1}, \mathbf{u}^{\text{w}}_{k})] &- r(\overline{\mathbf{x}}^n_{k+1}, \overline{\mathbf{u}}^n_{k})|  \\ &\leq\frac{L (Q+M + N)(1- \gamma_d^{K+1})}{1 - \gamma_d}.
\end{align*}
Therefore, we have:
\begin{align}
    |\mathbb{E}_{\pi, p^{\text{w}}} [\sum^{K-1}_{k=0} \gamma_d r(\mathbf{x}^{\text{w}}_{k+1}, \mathbf{u}^{\text{w}}_k) ] &- \sum^{K-1}_{k=0} \gamma_d r(\overline{\mathbf{x}}^n_{k+1}, \overline{\mathbf{u}}^n_k) | \nonumber  \\ &\leq \frac{L (Q+M + N)(1- \gamma_d^{K+1})}{1 - \gamma_d}, 
    \label{lr6}
\end{align}
and 
\begin{align} 
    |\mathbb{E}_{\pi, p^{\text{w}}} [\sum^{\infty}_{k=0} \gamma_d r(\mathbf{x}^{\text{w}}_{k+1}, \mathbf{u}^{\text{w}}_k) ] &- \sum^{\infty}_{k=0} \gamma_d r(\bar{\mathbf{x}}^n_{k+1}, \overline{\mathbf{u}}^n_k) | \nonumber  \\ &\leq \frac{L (Q+M + N)}{1 - \gamma_d},  0 \leq \gamma_d < 1.
    \label{lr6_}
\end{align}
Hence, the terms $\|\mathbf{T}^{\text{w}n}_z\|_{H_\infty} $ and $\| {\mathbf{\widetilde{K}}^f}_z\|_{H_\infty}$ directly control the magnitude of the upper impact of domain change on the expected reward.
Larger $\|\mathbf{T}^{\text{w}n}_z\|_{H_\infty} $ and $\| {\mathbf{\widetilde{K}}^f}_z\|_{H_\infty}$ lead to a higher impact, meaning more vulnerability to domain change.
Therefore, designing a DRL algorithm such that these terms have smaller values improves the robustness and domain generalization of the algorithm.$ \qed$
\vspace{3pt}

It is important to note that many nonlinear functions satisfy the Lipschitz condition if they are varied at a controlled rate. Typical examples include certain polynomial functions, bounded exponential functions, and sigmoid-like functions.
Moreover, for more general nonlinear functions $r(\mathbf{x}_{k+1}, \mathbf{u}_k)$, it is possible to use specific properties of the known function $r(\mathbf{x}_{k+1}, \mathbf{u}_k)$ to derive the upper limit on how domain changes affect the expected cumulative reward of a trained DRL.

Now, we want to drive the generalization error bound defined in (\ref{GE}) for the trained DRL algorithm based on \textbf{Theorem 2}, \textbf{Corollary 1}, and \textbf{Corollary 2}.
\vspace{0.2cm}

\noindent
\textbf{Assumption 6}: We assume that the expected deviation of \( (\mathbf{x}^n_{k+1}, \mathbf{u}^n_k) \) from its mean is bounded by constant \( C \):
\[
\mathbb{E}_{\pi, p^n} \left\| (x^n_{k+1}, u^n_k) - (\overline{x}^n_{k+1}, \overline{u}^n_k) \right\|_2 \leq C.
\]

\vspace{0.2cm}
\noindent
\textbf{Corollary 3}. Given a trained DRL policy, for any domain change such that $\|{\mathbf{w}}_z\|_{H_\infty} \leq \gamma$, the generalization error bound for the trained DRL algorithm is:\\
\begin{align*}
| \mathbb{E}_{\pi, p^{\text{w}}} [\sum^{\infty}_{k=0} \gamma_d^k r(\mathbf{x}^{\text{w}}_{k+1}, \mathbf{u}^{\text{w}}_k)] - \mathbb{E}_{\pi, p^n} [\sum^{\infty}_{k=0} \gamma_d^k r(\mathbf{x}^n_{k+1}, \mathbf{u}^n_k)] | \leq \nonumber\\\frac{L (Q+M+N) + L C}{1 - \gamma_d}.
\label{GE_C}
\end{align*}

\noindent
\textbf{Proof}.
First, we want to find a bound for the difference:
\begin{equation*}
   \left| r(\overline{\mathbf{x}}^n_{k+1}, \overline{\mathbf{u}}^n_k) - \mathbb{E}_{\pi, p^n} [r(\mathbf{x}^n_{k+1}, \mathbf{u}^n_k)] \right|. 
   \label{lr7}
\end{equation*}
As \( r(\mathbf{x}_{k+1}, \mathbf{u}_k) \) is Lipschitz continuous in both \( \mathbf{x} \) and \( \mathbf{u} \) with constant \( L \), we get:
\begin{align*}
| r(\overline{\mathbf{x}}^n_{k+1}, \overline{\mathbf{u}}^n_k) - r(\mathbf{x}^n_{k+1}, \mathbf{u}^n_k) | \leq  L || (\overline{\mathbf{x}}^n_{k+1}, \overline{\mathbf{u}}^n_k) - (\mathbf{x}^n_{k+1}, \mathbf{u}^n_k) ||_2.
\end{align*}
Taking expectation over \( \pi, p^n \) on both sides:
\begin{align*}
\mathbb{E}_{\pi, p^n} [| r(\overline{\mathbf{x}}^n_{k+1}, \overline{\mathbf{u}}^n_k) - &r(\mathbf{x}^n_{k+1}, \mathbf{u}^n_k)|]
\leq L \nonumber\\ &\mathbb{E}_{\pi, p^n} [|| (\overline{\mathbf{x}}^n_{k+1}, \overline{\mathbf{u}}^n_k) - (\mathbf{x}^n_{k+1}, \mathbf{u}^n_k) ||_2].
\end{align*}
Thus, based on \textbf{Assumption 6}, we have:
\[
\mathbb{E}_{\pi, p^n} [\left| r(\overline{\mathbf{x}}^n_{k+1}, \overline{\mathbf{u}}^n_k) - r(\mathbf{x}^n_{k+1}, \mathbf{u}^n_k) \right|] \leq L C.
\]
The absolute value function is convex, and by applying Jensen’s inequality, we have:
\[\left|\mathbb{E}_{\pi, p^n}[r(\overline{\mathbf{x}}^n_{k+1}, \overline{\mathbf{u}}^n_k) - r(\mathbf{x}^n_{k+1}, \mathbf{u}^n_k)]  \right| \leq L C,\]
\[\left| r(\overline{\mathbf{x}}^n_{k+1}, \overline{\mathbf{u}}^n_k) -\mathbb{E}_{\pi, p^n}[ r(\mathbf{x}^n_{k+1}, \mathbf{u}^n_k)]  \right| \leq L C.\]
Now, we sum over all \( k \) with discount factor \( \gamma_d^k \):
\begin{align}
\sum^{\infty}_{k=0} \gamma_d^k \left| r(\overline{\mathbf{x}}^n_{k+1}, \bar{\mathbf{u}}^n_k) -\mathbb{E}_{\pi, p^n}[ r(\mathbf{x}^n_{k+1}, \mathbf{u}^n_k)]  \right| \leq \frac{L C}{1 - \gamma_d}.
\label{lr8}
\end{align}
By using the triangle inequality:
\begin{align}
| \sum^{\infty}_{k=0} \gamma_d^k r(\overline{\mathbf{x}}^n_{k+1}, \overline{\mathbf{u}}^n_k) - \sum^{\infty}_{k=0} \gamma_d^k \mathbb{E}_{\pi, p^n} [ r(\mathbf{x}^n_{k+1}, \mathbf{u}^n_k)] | \leq \frac{L C}{1 - \gamma_d}.
\label{lr9}
\end{align}
Considering equation (\ref{lr6_}) and combining it with (\ref{lr9}), we get:
\begin{align*}
| \mathbb{E}_{\pi, p^{\text{w}}} [\sum^{\infty}_{k=0} \gamma_d^k r(\mathbf{x}^{\text{w}}_{k+1}, \mathbf{u}^{\text{w}}_k)] - \mathbb{E}_{\pi, p^n} [\sum^{\infty}_{k=0} \gamma_d^k r(\mathbf{x}^n_{k+1}, \mathbf{u}^n_k)] | \leq \nonumber\\\frac{L (Q+M+N) + L C}{1 - \gamma_d}.\qed
\label{lr8}
\end{align*}
Thus, it is worth emphasizing that larger values of $\|\mathbf{T}^{\text{w}n}_z\|_{H_\infty} $ and $\| {\mathbf{\widetilde{K}}^f}_z\|_{H_\infty}$ lead to a higher generalization error bound.
\vspace{-0.3cm}
\section{Experiments in a wireless communication
environment}
In this section, we demonstrate the applicability of the proposed generalization analysis in a wireless communication environment. Specifically, we focus on UAV trajectory design in a UAV-assisted millimeter-wave (mmWave) network. 
This application introduces real-world constraints and challenges, such as dynamic channel conditions and user mobility, which are essential for assessing the robustness of DRL algorithms in terms of generalization. We first present the system model and problem formulation, followed by solutions using two DRL algorithms: soft actor-critic (SAC) and proximal policy optimization (PPO). Next, we evaluate the generalizability of these DRL algorithms using the proposed analysis framework. It is important to note that our objective is to validate the theoretical framework rather than to develop a new DRL algorithm. 
\vspace{0.1cm}

\noindent
\textbf{System Model and Assumptions} 
We consider a UAV-assisted wireless network consisting of $J$ mobile ground users (GUs). Initially, both the UAV and mobile GUs are randomly distributed across a service area of $A=A_1*A_2$. The set of mobile GUs is represented by $\mathcal{J} = \{0,1,\ldots,J-1\}$. The system is analyzed over multiple time intervals, with each interval evenly divided into $K$ time steps of duration $\kappa$, normalized to one. The UAV provides downlink communication for mobile GUs in mmWave frequency bands. 
The operational range of mmWave-enabled UAVs is limited due to the short propagation distance of mmWave under atmospheric conditions. To address this, the UAV's mission is to navigate autonomously toward the GUs and maximize the downlink coverage for the mobile GUs within its coverage area. Specifically, the objectives are to optimize the downlink coverage for mobile GUs, ensuring fairness through the UAV trajectory design. 
We adopt the following motion model for GUs:
\begin{align}
    v^j_k &= h_1 v^j_{k - 1} + (1 - h_1)\bar{v} + \nu_k, \\
    \phi^j_k &= \phi^j_{k - 1} + h_2 \bar{\phi},
\end{align}
\noindent where $\bar{v}$ represents the average speed, $\nu$ accounts for random uncertainty in speed, and $\bar{\phi}$ is the average steering angle, $0 \leq h_1, h_2 \leq 1$ are parameters that control the influence of the previous state. In addition, $h_2$ follows an $\epsilon$-greedy strategy, where the GU maintains its current direction with a probability of $\epsilon$ or selects a random direction otherwise. At time $k \in \{0, 1, \ldots, K - 1\}$, the UAV's position is \(\mathbf{p}^\text{UAV}_k = (x^\text{UAV}_k, y^\text{UAV}_k, H)\),
where $H$ is the constant altitude of the UAV. The horizontal projection of the UAV’s position is represented as \(
\hat{\mathbf{p}}^\text{UAV}_k = (x^\text{UAV}_k, y^\text{UAV}_k),
\) and its path over time is described by $\{\hat{\mathbf{p}}^\text{UAV}_k\}$. The position of the $j$-th GU is \(\mathbf{p}^j_k = (x^j_k, y^j_k, 0)\). The UAV’s movement is constrained by its maximum speed $V_\text{max}^\text{UAV}$ and the time interval $\kappa$ between steps. This ensures that:
\begin{equation}
    \|\hat{\mathbf{p}}^\text{UAV}_k - \hat{\mathbf{p}}^\text{UAV}_{k-1}\|_2 \leq \kappa V_\text{UAV}^\text{max}, \quad \forall k\in \{0, 1, \ldots, K - 1\}.
\end{equation}
High-frequency bands, such as mmWave, exhibit limited scattering capability, resulting in the channel being largely governed by the line-of-sight (LoS) path. Therefore, Non-line-of-sight (NLoS) transmissions are considered negligible because of the substantial molecular absorption. The path-loss coefficient \(h_{g}^j\) for GU $j$, described as \(h_{g}^j = h_{gp}^j h_{ga}^j\), where \(h_{gp}^j\) accounts for propagation loss and \(h_{ga,j}\) represents molecular absorption \cite{chang2022integrated}. The propagation loss is \(h_{gp}^j = \frac{c \sqrt{G^\text{UAV} G^j}}{4\pi f^j d^j},\) with \(G^\text{UAV}\) and \(G^j\) being the transmission and reception gains, \(c\) as the speed of light, \(f^j\) is the operational frequency used for GU $j$, and \(d^j\) the distance between the UAV and GU $j$. The molecular absorption coefficient is defined as \(h_{ga}^j = e^{-\frac{1}{2} \alpha(f^j) d^j}\), where \(\alpha(f^j)\) is the medium absorption factor which depends on the amount of water vapor molecules present and the operating mmWave frequency being used. Accordingly, the downlink transmission rate from the UAV at GU $j$ in bits per second is given by \cite{chang2022integrated}: 
\begin{equation}
    R^j= \omega \log_2 \left(1 + \frac{P|h_{g}^j|^2}{N_0}\right),
\end{equation}
where \(\omega\) denotes the bandwidth allocated to GU \(j\), $P$ is the constant value of power, and $N_0$ is noise power. For every GU \(j \in \mathcal{J}\), it is assumed that a minimum downlink transmission rate, represented by \(R^j \geq R^{\text{min}}\), must be maintained to meet its quality of service (QoS) requirements. Notably,  each GU does not require continuous data transmission, but must meet the minimum data rate whenever it is actively being served. Additionally, the
parameters of $h_{g}^j$ is considered as specified in \cite{chang2022integrated}.
\vspace{0.1cm}

\noindent
\textbf{Problem Formulation}: 
The UAV trajectory problem is formulated as follows:
\begin{equation}
\begin{aligned}
&\max_{\{\hat{\mathbf{p}}^{\text{UAV}}_k\}}\sum_{k=0}^{K-1}\left(a \frac{\sum_{j=0}^{J-1}s^j_ k}{J} + (1-a) I^{\text{fairness}}_k\right) \\
&\text{subject to}:\\ 
&C_{1}: \text{equations (36) and (37)},\\
&C_{2}: \text{equation (38)}, \\
&C_{3}: R^{j}_k \geq s^j_k {R^{\text{min}}},\\
&C_{4}: d^{j}_k s^j_k \leq D^{\text{max}}_{\text{UAV}},\\
\end{aligned}
\label{optequ} 
\end{equation}
where \(s^j_k\) represents the indicator function showing whether GU \(j\) is being served by the UAV at time step \(k\). Specifically, \(s^j_k = 1\) indicates that GU \(j\) is being served, and \(s^j_k = 0\) otherwise. In addition, \(I^{\text{fairness}}\) is Jain's fairness index, defined as \(I^{\text{fairness}}_k = \frac{\left(\sum_{j=0}^{J-1} s^j_k\right)^2}{J^2 \sum_{i=0}^{J-1} (s^j_k)^2}\), \(0 \leq a \leq 1\) represents the priority given to optimizing both the number of served GUs and the fairness. Furthermore, \(C_1\) and \(C_2\) denote the movement model of the GUs and the UAV's maximum speed limitation, respectively. \(C_3\) captures the QoS requirements for the served GUs, and \(C_4\) indicates the operational coverage limit of the mmWave-enabled UAV.
\vspace{0.1cm}

\noindent
\textbf{Proposed Solution}:
The non-convexity of problem (\ref{optequ}) arises from non-linear terms, such as Jain's fairness index. Moreover, the inclusion of random variables adds complexity and uncertainty to the optimization. To tackle this problem, we propose employing DRL algorithms (SAC and PPO) which are well-suited for solving non-convex problems in wireless applications. 
The state vector for both DRL algorithms is defined by $\mathbf{x}_k = ( \mathbf{p}^j_k,\hat{\mathbf{p}}^{\text{UAV}}_k)$ and the action is defined as $\mathbf{u}_k= \hat{\mathbf{p}}^{\text{UAV}}_{k+1}$. Additionally, the reward function is considered as $r(\mathbf{x}_{k+1},\mathbf{u}_k)= a \frac{\sum_{j=0}^{J-1}s^j_ k}{J} + (1-a) I^{\text{fairness}}_k+\beta \Delta_k$ , where $\Delta_k$ denotes whether the UAV violates the speed limitation. $\Delta_k = 1$ if the UAV violates the speed limitation, otherwise, $\Delta_k = 0$.
\begin{table}[t!]
\centering
\caption{Simulation Parameters}
\begin{tabular}{c|c}  
    \hline\hline
    \rowcolor{gray!20} \textbf{Parameter} & \textbf{Value} \\ 
    \hline
    Service area  ($A_1 \times A_2$) & $100 \times 100$ m$^2$ \\
    \hline
    \rowcolor{gray!15} Number of GUs  ($J$)  & 20 \\
    \hline
    UAV height  ($H$)  & 30 m \\
    \hline
    \rowcolor{gray!15} Time step length ($\kappa$)  & 0.1 s \\ 
    \hline
    UAV’s max speed ($V_{\text{UAV}}^{\max}$) & 30 m/s \\
    \hline
    \rowcolor{gray!15} UAV coverage area & 50 m \\
    \hline
    GU's average speed ($\bar{v}$) & 3 m/s \\
    \hline
    \rowcolor{gray!15} GU speed uncertainty ($\nu$) & 0.5 to 0.8 \\ 
    \hline
    Greedy strategy for GU direction ($\epsilon$) & 0.5 to 0.8  \\
    \hline
    \rowcolor{gray!15} Total mmWave bandwidth & 400 MHz \\
    \hline
    Transmit power (P)  & 0.2512 Watt \\
    \hline
    \rowcolor{gray!15} Central frequency  & 30 GHz \\
    \hline
    Noise power ($N_0$)  & -85 dBm \\
    \hline
    \rowcolor{gray!15} Minimum rate (${R}^{\text{min}}$)  & 150 Mb/s \\
    \hline\hline
    \rowcolor{gray!20} \textbf{DRL} & \textbf{SAC / PPO}\\
    \hline
    Number of layers & 4 / 5 \\
    \hline
    \rowcolor{gray!15} Nodes per layer & 256, 256 / 64, 64, 8\\
    \hline
    Reward scale & 4 / - \\
    \hline
    \rowcolor{gray!15} Learning rate & 0.0003 / 0.007-0.01  \\
    \hline
    Discount factor & 0.9 / 0.99 \\
    \hline
    \rowcolor{gray!15} Clipping hyper-parameter & - / 0.2 \\
    \hline
     Entropy coefficient  & - / 0.5 \\
    \hline
\end{tabular}
\label{tab: Table1}
\end{table} 

\begin{figure}[t!]
    \centering
    \includegraphics[width=3.5in]{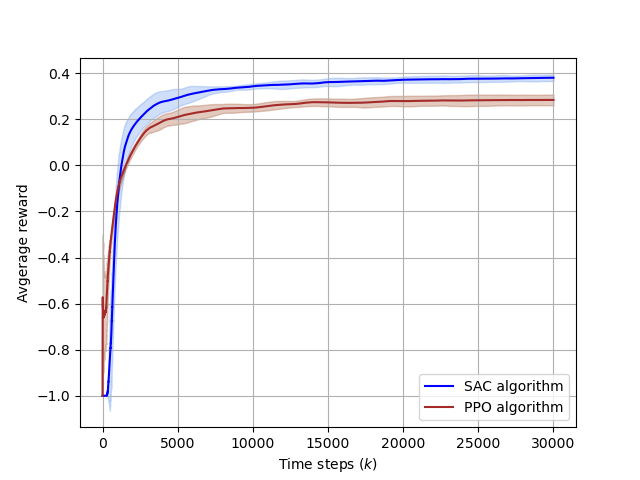}
    \caption{Training reward convergence of SAC and PPO in UAV trajectory with 95\% confidence intervals.}
    \label{reward_uav}
\end{figure}

\noindent
\textbf{Numerical Results}
The parameters of the simulated system model are detailed in Table \ref{tab: Table1}. The system is tested over several runs, where each run includes multiple episodes. Each episode is divided into $K$ time steps of length $\kappa$, normalized to one. The DRL-related parameters used in the simulations are also provided in Table~\ref{tab: Table1}. Fig. \ref{reward_uav} illustrates the reward convergence curve during training. The simulations are conducted over four runs, with a 95\% confidence interval. Fig. \ref{reward_uav} shows that SAC achieves higher reward values compared to PPO. Although PPO converges to lower reward values, both algorithms exhibit similar variability across simulation runs, indicating comparable robustness to uncertainties in the training setup. These uncertainties include random variations in speed and noise power, which do not change the domain (i.e., the conditional transition probability function of the environment).
\begin{table}[t!]
\centering
\caption{$H_{\infty}$ norms for SAC and PPO Algorithms}
\renewcommand{\arraystretch}{1.1}
\setlength{\tabcolsep}{6pt} 

\begin{tabular}{lcc}
\toprule
\rowcolor{gray!15} \textbf{DRL Algorithm} & $\| \mathbf{T}^{\text{w}n}_z \|_{\text{H}_\infty}$ & $\| \widetilde{\mathbf{K}}^f_z \|_{\text{H}_\infty}$ \\
\midrule
SAC & 2602.17 & 0.3519 \\
\rowcolor{gray!15} PPO & 5933.16 & 0.3764 \\
\bottomrule
\end{tabular}
\label{tab: Table2}
\end{table}

Next, considering the trained SAC and PPO algorithms,  we employ PyDMD \cite{demo2018pydmd, ichinaga2024pydmd} (a Python package designed for DMD) to compute $\widetilde{\mathbf{K}}^h$ from equation (\ref{inter_general state}). Notably, since the dimensions of $\mathbf{u}_{k}$ and $\mathbf{x}_{k}$ do not match, step 3 of the exact DMD algorithm cannot be applied to compute $\widetilde{\mathbf{K}}^f$ of equations (\ref{inter_u}). As a result, we adopt an SVD-based approach to approximate the non-square linear operator $\widetilde{\mathbf{K}}^f$, utilizing the Moore–Penrose pseudoinverse of $\mathbf{x}_{k}$. 
The computation of $\widetilde{\mathbf{K}}^h$ and $\widetilde{\mathbf{K}}^f$ is performed using data that is collected by running the trained SAC and PPO models under conditions that are similar to those used during training. The data is collected across multiple independent runs and $K =$ 30{,}000 time steps. Subsequently, we calculate $\|\mathbf{T}^{wn}_z\|_{H_\infty}$ and $\| {\widetilde{\mathbf{K}}^f}_z\|_{H_\infty}$ as presented in Table~\ref{tab: Table2}. The value of $\| \mathbf{T}^{\text{w}n}_z \|_{\text{H}_\infty}$ and $\| {\widetilde{\mathbf{K}}^f}_z\|_{H_\infty}$ for the SAC algorithm are lower than those for the PPO algorithm. As suggested by \textbf{Corollary 2}, this implies that the maximum impact of domain change on SAC's performance will be lower than on PPO's. This will be confirmed in the subsequent experimental results.

\begin{figure}
    \centering
    \includegraphics[width=3in]{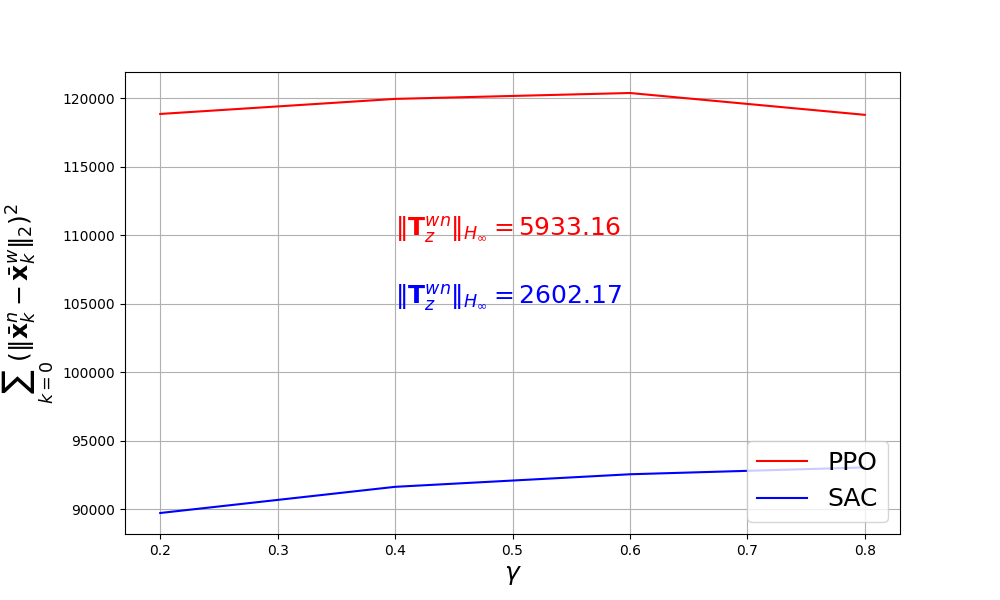}
    \caption{Impact of domain changes on states: SAC vs. PPO}
    \label{state_sac_ppo}
\end{figure}

\begin{figure}
    \centering
    \includegraphics[width=3in]{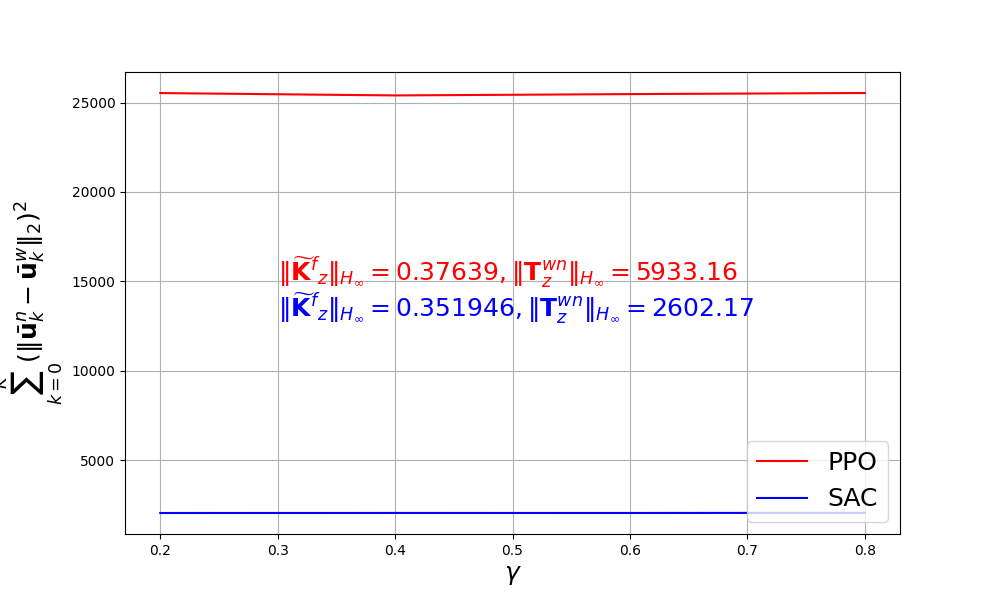}
    \caption{Impact of domain changes on actions: SAC vs. PPO}
    \label{action_sac_ppo}
\end{figure}

To introduce domain changes in UAV trajectory environment, we adjust three factors: the average speed of mobile GUs ($\bar{v}$), noise power ($N_0$), and the medium absorption factor (\(\alpha(f_j)\)). For each factor, we add random values sampled from normal distributions. The means of these distributions are proportional to $\gamma$, calculated as: $\gamma \times \text{the value of that factor}$. Figures \ref{state_sac_ppo} and \ref{action_sac_ppo} confirm that the maximum impact of domain changes (characterized by $\gamma$) on the states and actions is primarily governed by the term \( \|\mathbf{T}^{\text{w}n}_z\|_{{H}_\infty} \) for the states, and both \( \|\mathbf{T}^{\text{w}n}_z\|_{{H}_\infty} \) and \( \| \widetilde{\mathbf{K}}^f_z\|_{{H}_\infty} \) for the actions, as established in \textbf{Theorem 2} and \textbf{Corollary 1}.

\begin{figure*}[t!]
  \centering
  \begin{minipage}{0.497\textwidth}
    \centering
    \includegraphics[width=\textwidth]{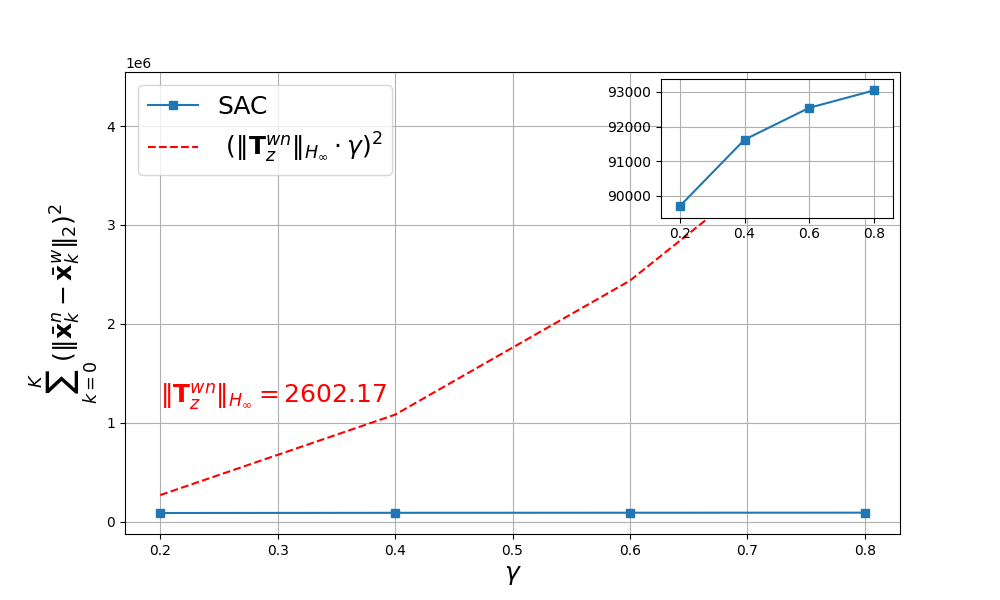}
    (a) 
    \label{state_sac_uav}
  \end{minipage}
  \hfill
  \begin{minipage}{0.497\textwidth}
    \centering
    \includegraphics[width=\textwidth]{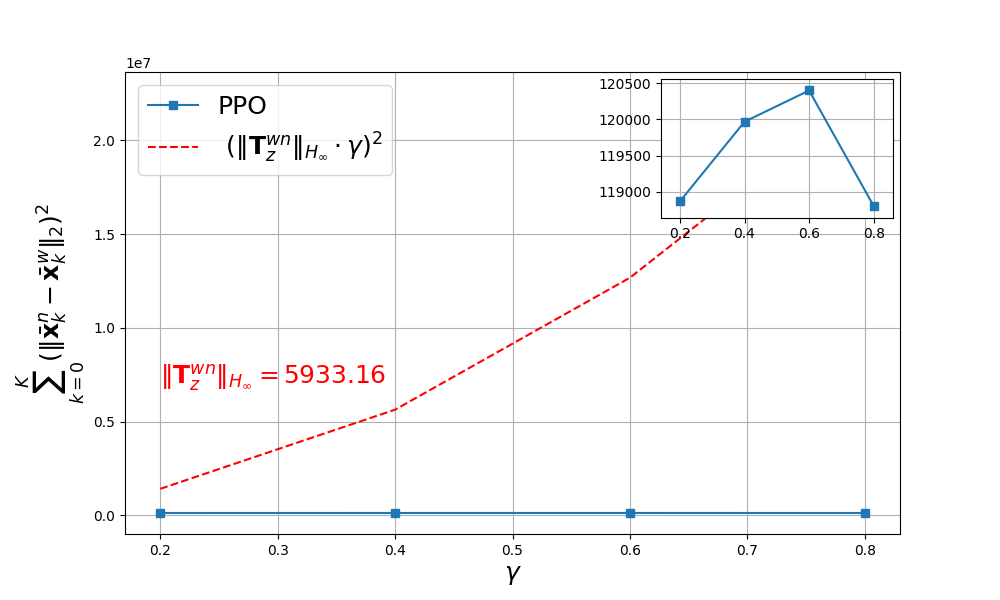}
    (b) 
    \label{state_ppo_uav}
  \end{minipage}
  \begin{minipage}{0.497\textwidth}
    \centering
    \includegraphics[width=\textwidth]{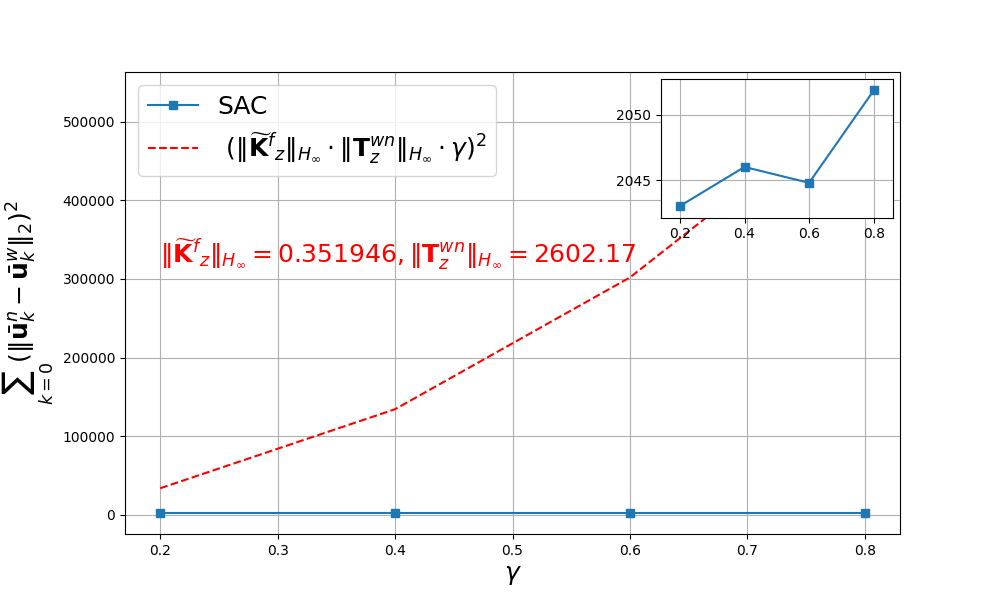}
    (c)  
    \label{action_sac_uav}
  \end{minipage}
  \hfill
  \begin{minipage}{0.497\textwidth}
    \centering
    \includegraphics[width=\textwidth]{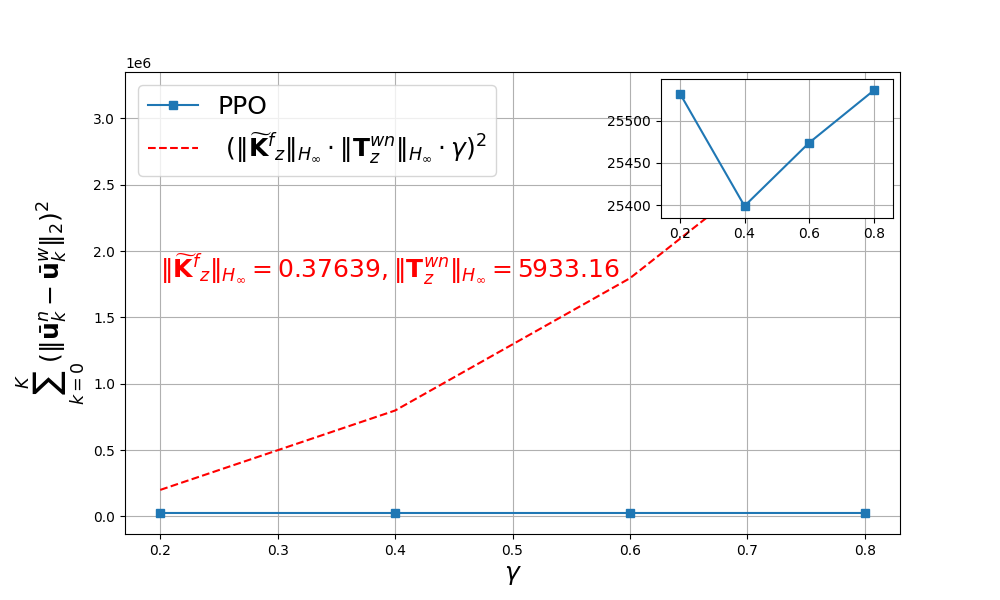}
    (d)  
    \label{action_ppo_uav}
  \end{minipage}
  \caption{ Effect of domain changes. (a) Effect of domain changes on SAC states. (b) Effect of domain changes on PPO states. (c) Effect of domain changes on SAC action. (d) Effect of domain changes on PPO action.}
  \label{state_action_uav}
\end{figure*}

Fig. \ref{state_action_uav} further investigates how tight or loose the upper bounds are on the maximum effect of domain changes on states and actions, as derived in \textbf{Theorem 2} and \textbf{Corollary 1}. Although these bounds are validated, it's important to emphasize that the use of the $H_{\infty}$ norm leads to conservative, worst-case estimates, which is reflected in the figure. Despite the conservative nature of the estimate, the bounds provide meaningful insights into the generalization behavior of DRL algorithms, in line with the conclusions of \textbf{Corollary 2}.

\begin{figure}[h!] 
    \centering
    \includegraphics[width=3.5in]{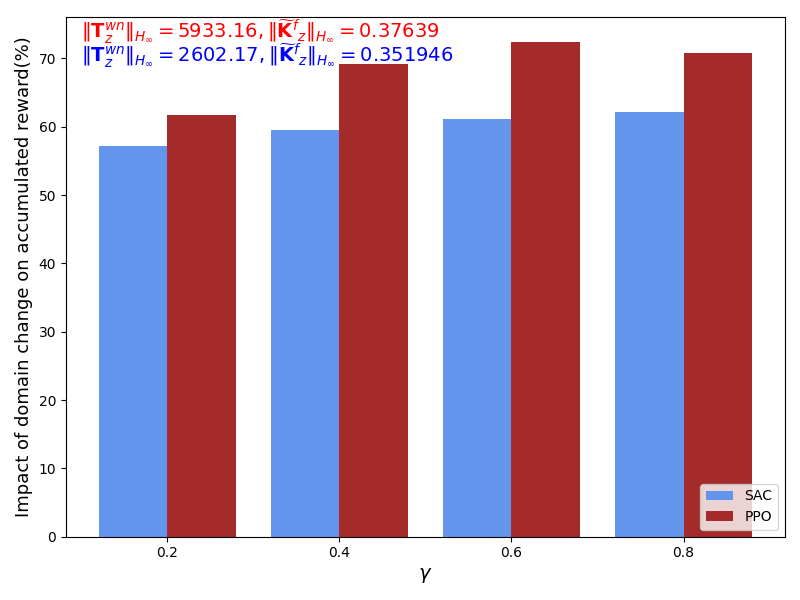}
    \caption{Percentage of the average impact of domain change on reward in SAC vs. PPO algorithms}
    \label{reward_change_uav}
\end{figure}

As discussed in Section IV.B, the terms $\|\mathbf{T}^{wn}_z\|_{H_\infty}$ and $\| {\widetilde{\mathbf{K}}^f}_z\|_{H_\infty}$ directly control
the magnitude of the upper impact of domain change on
the expected reward. This relationship is validated in  Fig. \ref{reward_change_uav}. It confirms that domain changes have a notably greater impact on the accumulated reward of the PPO algorithm compared to SAC. This difference is due to the much larger value of $\|\mathbf{T}^{\text{w}n}_z\|_{H_\infty}$ in PPO than in SAC.
\section{Conclusion}
For DRL algorithms, we have developed a novel analytical framework for generalizability analysis under domain changes. To understand how domain changes affect DRL performance, we have analyzed the evolution of internal variables. Specifically, we have explored how states and actions evolve over time steps under such changes for a trained DRL policy. More specifically, we have introduced interpretable representations of the state and action dynamics by employing Koopman operator theory and the DMD method. Then we have applied the H$_\infty$ norm to quantify the maximum impact of domain changes on the DRL reward function using the most dominant eigenvalues of the underlying dynamics.  Next, we have applied the proposed framework to assess the generalizability of several DRL algorithms in a wireless communication scenario.  

A key focus of our framework is the analysis of the most dominant eigenvalues of the underlying dynamics, as they significantly influence the sensitivity of internal variables to domain changes. In this work, we estimate these eigenvalues by applying basic DMD to the Koopman operator that tracks the expected values of the system states and actions. Although basic DMD is often sufficient to extract these critical eigenvalues with high probability, achieving tighter generalization bounds requires more accurate interpretable models that better capture the underlying dynamics. To this end, we plan to enhance our analysis by employing a Koopman observer capable of tracking not only the expected values of system states and actions,
$\overline{\mathbf{x}}_{k}$ and $\overline{\mathbf{u}}_{k}$, 
but also their associated covariances.
\balance
\bibliographystyle{ieeetr}
\bibliography{ref}
\end{document}